\documentclass[11pt]{article}

\usepackage{fullpage}
\usepackage[ruled, linesnumbered, vlined, commentsnumbered]{algorithm2e}
\usepackage{graphicx,subfig} % figure related
\usepackage{amsfonts,amssymb,amsmath,amsthm,amsopn,mathtools}	% math related
\usepackage{booktabs,diagbox,colortbl,multirow,tabularx,threeparttable,hhline}
\usepackage[listings,skins,breakable]{tcolorbox}

\usepackage{enumerate}
\usepackage{authblk}
\usepackage{footnote}
\usepackage{hyperref}
\usepackage{prettyref}
\usepackage{cite}
\usepackage{setspace}
\usepackage{color}
\usepackage{xcolor}  % Required for custom colors
\usepackage{scrpage2}
\usepackage{geometry}

\usepackage{tikz}
\usepackage{pgfplots}
\usetikzlibrary{positioning,shapes,shadows,arrows,calc}
\tikzstyle{component}=[rectangle, draw=black, rounded corners, fill=blue!40, drop shadow, text centered, anchor=north, text=white, minimum height=1cm]
\tikzstyle{arrow}=[->, thick]

\pgfplotsset{compat=1.12}
\usetikzlibrary{intersections}
\usetikzlibrary{pgfplots.statistics}
\usepgfplotslibrary{fillbetween}

\definecolor{myblue}{RGB}{34,31,217}
\definecolor{mycyan}{gray}{.7}
\definecolor{Gray}{gray}{0.9}

\newtheorem{remark}{Remark}

\newcommand{\pref}{\prettyref}

\newrefformat{fig}{Fig.~\ref{#1}}
\newrefformat{tab}{Table~\ref{#1}}
\newrefformat{sec}{Section~\ref{#1}}
\newrefformat{alg}{Algorithm~\ref{#1}}
\newrefformat{property}{Property~\ref{#1}}
\newrefformat{theorem}{Theorem~\ref{#1}}
\newrefformat{definition}{Definition~\ref{#1}}
\newrefformat{corollary}{Corollary~\ref{#1}}
\newrefformat{lemma}{Lemma~\ref{#1}}
\newrefformat{conj}{Conjecture~\ref{#1}}
\newrefformat{def}{Definition~\ref{#1}}
\newrefformat{eq}{equation~(\ref{#1})}
\newrefformat{app}{Appendix~\ref{#1}}

\usepackage{lscape}

\begin{document}

%% title
\title{\vspace{-1ex}\LARGE\textbf{Do We Really Need to Use Constraint Violation in Constrained Evolutionary Multi-Objective Optimization?}~\footnote{This manuscript is submitted for potential publication. Reviewers can use this version in peer review.}}

%% authors and affiliations
\author[1]{\normalsize Shuang Li}
\author[2]{\normalsize Ke Li}
\author[1]{\normalsize Wei Li}
\affil[1]{\normalsize Control and Simulation Center, Harbin Institute of Technology, China}
\affil[2]{\normalsize Department of Computer Science, University of Exeter, EX4 4QF, Exeter, UK}
\affil[$\ast$]{\normalsize Email: \texttt{k.li@exeter.ac.uk}}

\date{}
\maketitle

\vspace{-3ex}
{\normalsize\textbf{Abstract: } }Constraint violation has been a building block to design evolutionary multi-objective optimization algorithms for solving constrained multi-objective optimization problems. However, it is not uncommon that the constraint violation is hardly approachable in real-world black-box optimization scenarios. It is unclear that whether the existing constrained evolutionary multi-objective optimization algorithms, whose environmental selection mechanism are built upon the constraint violation, can still work or not when the formulations of the constraint functions are unknown. Bearing this consideration in mind, this paper picks up four widely used constrained evolutionary multi-objective optimization algorithms as the baseline and develop the corresponding variants that replace the constraint violation by a crisp value. From our experiments on both synthetic and real-world benchmark test problems, we find that the performance of the selected algorithms have not been significantly influenced when the constraint violation is not used to guide the environmental selection.


%!TeX root=main.tex

\section{Introduction}
\label{sec:introduction}

%For example, in portfolio optimization~\cite{ArnoneLT93}, minimizing the risk and maximizing the return are two typical objectives. In the meanwhile, a typical constraint is the sum of the invested amounts must exactly meet the available capital.
Real-world optimization problems in science~\cite{ThurstonDSS03}, engineering~\cite{Andersson03} and economics~\cite{PonsichJC13} usually involve multiple conflicting objectives under a number of equality and inequality constraints, a.k.a. constrained multi-objective optimization problems (CMOPs). In this paper, we consider the CMOP defined as follows:
\begin{equation}
    \begin{array}{l l}
        \mathrm{minimize} \quad\, \mathbf{F}(\mathbf{x})=(f_{1}(\mathbf{x}),\cdots,f_{m}(\mathbf{x}))^{T}\\
        \mathrm{subject\ to} \;\;\, \mathbf{g}(\mathbf{x})\leq 0\\
        \mathrm{\ } \quad\quad\quad\quad\;\;\, \mathbf{h}(\mathbf{x})=0\\
        \mathrm{\ } \quad\quad\quad\quad\;\;\, \mathbf{x}=(x_1,\cdots,x_n)^T\in\Omega
    \end{array},
    \label{MOP}
\end{equation}
where $\Omega=[x_i^L,x_i^U]^n_{i=1}\subseteq\mathbb{R}^n$ defines the search (or decision variable) space and $\mathbf{x}$ is an $n$-dimensional vector therein. $\mathbf{F}:\Omega\rightarrow\mathbb{R}^m$ constitutes $m$ conflicting objective functions, and $\mathbb{R}^m$ is the objective space. $\mathbf{g}(\mathbf{x})=(g_1(\mathbf{x}),\cdots,g_p(\mathbf{x}))^T$ and $\mathbf{h}(\mathbf{x})=(h_1(\mathbf{x}),\cdots,h_q(\mathbf{x}))^T$ are vectors of inequality and equality constraints respectively. Given a CMOP, the degree of constraint violation of a solution $\mathbf{x}$ at the $j$-th constraint is calculated as:
\begin{equation}
    c_i(\mathbf{x})=
    \begin{cases}
        \langle g_j(\mathbf{x})/a_j-1\rangle, & j=1,\cdots,q,\ i=j, \\
        \langle |h_k(\mathbf{x})/b_k-1|-\epsilon\rangle, & k=1,\cdots,p,\ i=k+q,
    \end{cases}
\end{equation}
where $\epsilon$ is a small tolerance term (e.g., $\epsilon=10^{-6}$) that relaxes the equality constraints to the inequality constraints. $a_j$ and $b_k$ where $j\in\{1,\cdots,q\}$ and $k\in\{1,\cdots,p\}$ are normalization factors of the corresponding constraints. $\langle\alpha\rangle$ returns 0 if $\alpha\geq 0$ otherwise it returns the negative of $\alpha$. Given a CMOP, the constraint violation (CV) value of a solution $\mathbf{x}$ is calculated as:
    \begin{equation}
        CV(\mathbf{x})=\sum_{i=1}^{\ell}c_i(\mathbf{x}),
        \label{eq:cv}
    \end{equation}
    where $\ell=p+q$. $\mathbf{x}$ is feasible in case $CV(\mathbf{x})=0$; otherwise $\mathbf{x}$ is infeasible. Given two feasible solutions $\mathbf{x}^1$ and $\mathbf{x}^2$, $\mathbf{x}^1$ is said to \textit{Pareto dominate} $\mathbf{x}^2$ (denoted as $\mathbf{x}^1\preceq\mathbf{x}^2$) if and only if $f_i(\mathbf{x}^1)\leq f_i(\mathbf{x}^2)$, $\forall i\in\{1,\cdots,m\}$ and $\exists j\in\{1,\cdots,m\}$ such that $f_i(\mathbf{x}^1)<f_i(\mathbf{x}^2)$. A solution $\mathbf{x}^\ast\in\Omega$ is \textit{Pareto-optimal} with respect to (\ref{MOP}) if $\exists\mathbf{x}\in\Omega$ such that $\mathbf{x}\preceq\mathbf{x}^{\ast}$. The set of all Pareto-optimal solutions is called the \textit{Pareto-optimal set} (PS). Accordingly, $PF=\{\mathbf{F}(\mathbf{x})|\mathbf{x}\in PS\}$ is called the \textit{Pareto-optimal front} (PF).

Due to the population-based property, evolutionary algorithms (EAs) have been widely recognized as an effective approach for multi-objective optimization. Over the past three decades, much effort have been devoted to developing evolutionary multi-objective optimization (EMO) algorithms, e.g. elitist non-dominated sorting genetic algorithm (NSGA-II)~\cite{DebAPM02}, indicator-based EA (IBEA)~\cite{ZitzlerK04} and multi-objective EA based on decomposition (MOEA/D)~\cite{ZhangL07}. However, they cannot be directly applied to CMOPs without the assistance of a constraint handling technique (CHT), which can be seen as a selection mechanism to deal with constraints. In the 90s, some early endeavors to the development of EAs for solving CMOPs (e.g.,~\cite{FonsecaF98} and~\cite{CoelloC99}) are simply driven by a prioritization of the search for feasible solutions over \lq optimal\rq\ one. However, such methods are notorious for the loss of selection pressure in case the population is filled with infeasible solutions.

After the development of the constrained dominance relation~\cite{DebAPM02}, most, if not all, prevalent CHTs in the EMO community directly or indirectly depend on the CV defined in~\pref{eq:cv}. Specifically, a solution $\mathbf{x}^1$ is said to constraint-dominate $\mathbf{x}^2$, if: 1) $\mathbf{x}^1$ is feasible while $\mathbf{x}^2$ is not; 2) both of them are infeasible and $CV(\mathbf{x}^1)<CV(\mathbf{x}^2)$; or 3) both of them are feasible and $\mathbf{x}^1\prec\mathbf{x}^2$. By replacing the Pareto dominance relation with this constrained dominance relation, the state-of-the-art NSGA-II and NSGA-III~\cite{JainD14} can be readily used to tackle CMOPs. Borrowing this idea, several MOEA/D variants (e.g.,~\cite{JanZ10,JainD14,ChengJOS16,LiuWH20}) use the CV as an alternative criterion in the subproblem update procedure. Moreover, the constrained dominance relation is augmented with terms such as the number of violated constraints~\cite{OyamaSF07}, $\epsilon$-constraint~\cite{TakahamaS12,MartinezC14,AsafuddoulaRS15} and angle between each other~\cite{FanLCHLL16} to provide an additional selection pressure to infeasible solutions whose CV values have a marginal difference.

In addition to the above feasibility-driven CHTs, the second category aims at balancing the trade-off between convergence and feasibility during the search process. For example, Jim\'enez et al.~\cite{JimenezGSD02} proposed a min-max formulation that drives feasible and infeasible solutions evolve towards optimality and feasibility, respectively. In~\cite{RayTS01}, a Ray-Tai-Seow algorithm was proposed to simultaneously take the objective values, the CV along with the combination of them into consideration to compare and rank non-dominated solutions. Based on the similar rigour, some modified ranking mechanisms (e.g.,~\cite{Young05,AngantyrAA03,WoldesenbetYT09}) were developed by leveraging the information from both the objective and constraint spaces. Instead of prioritizing feasible solutions, some researchers (e.g.,~\cite{LiDZK15,PengLG17,SorkhabiAK17}) proposed to exploit information from infeasible solutions in case they can provide additional diversity to the current evolutionary population.

As a step further, the third category seeks to strike a balance among convergence, diversity and feasibility simultaneously. As a pioneer along this line, Li et al. proposed a two-archive EA that maintains two co-evolving and complementary populations to solve CMOPs~\cite{LiCFY19}. Specifically, one archive, denoted as the convergence-oriented archive (CA), pushes the population towards the PF; while the other one, denoted as the diversity-oriented archive, provides additional diversity. To complement the behavior of the CA, the DA explores the areas under-exploited by the CA including the infeasible region(s). In addition, to take advantage of the complementary effects of both CA and DA, a restricted mating selection mechanism was proposed to adaptively choose appropriate mating parents according to the evolution status of the CA and DA respectively. After~\cite{LiCFY19}, there have been a spike of efforts on the development of multi-population strategies (e.g.,~\cite{ShanL21,TianZXZJ21,LiuWT21,WangLZZG21,MingGWG22,MingGWL22}) to leverage some complementary effects of both feasible and infeasible solutions simultaneously for solving CMOPs.

Instead of the environmental selection, the last category tries to repair the infeasible solutions in order to drives them towards the feasible region(s). For example, a so-called Pareto descent repair operator~\cite{HaradaSOK06} was proposed to explore possible feasible solutions along the gradient information around infeasible solutions in the constraint space. In~\cite{JiaoLSL14}, a feasible-guided strategy was developed to guide infeasible solutions towards the feasible region along the \lq feasible direction\rq, i.e., a vector starting from an infeasible solution and ending up with its nearest feasible solution. In~\cite{SinghRS10}, a simulated annealing was applied to accelerate the progress of movements from infeasible solutions toward feasible ones.

\begin{remark}
    As discussed at the outset of this subsection, all these prevalent CHTs require the access of the CV. This applies to the last category, since it needs to access the gradient information of the CV. The implicit assumption behind the prevalent CHTs is the access of the closed form of the constraint function(s). However, this is not practical in the real world as problems are usually as a black box. In other words, we can only know that whether a solution is feasible or not.
\end{remark}

Bearing this consideration in mind, we come up with the overarching research question of this paper: \textit{do the prevalent CHTs in the EMO literature still work when we do not have an access to the CV?}

The rest of this paper is organized as follows. The experimental settings are summarized in~\pref{sec:settings} and the results are presented and analyzed in~\pref{sec:results}. Finally, \pref{sec:conclusion} concludes this paper and sheds some lights on future directions.

\section{Experimental Settings}
\label{sec:settings}

In this section, we introduce the experimental settings of our empirical study including the benchmark test problems, the peer algorithms, the performance metrics and statistical tests.

\subsection{Benchmark Test Problems}
\label{sec:benchmark}

In our empirical study, we pick up $45$ benchmark test problems widely studied in the literature to constitute our benchmark suite. More specifically, it consists of C1-DTLZ1, C1-DTLZ3, C2-DTLZ2 and C3-DTLZ4 from the C-DTLZ benchmark suite~\cite{JainD14}; DC1-DTLZ1, DC1-DTLZ3, DC2-DTLZ1, DC2-DTLZ3, DC3-DTLZ1, DC3-DTLZ3 chosen from the DC-DTLZ benchmark suite~\cite{LiCFY19}; and other $35$ problems picked up from the real-world constrained multi-objective problems (RWCMOPs) benchmark suite~\cite{KumarWALMSD21}. In particular, the RWCMOPs are derived from the mechanical design problems (denoted as RCM1 to RCM21), the chemical engineering problems (denoted as RCM22 to RCM24), the process design and synthesis problems (denoted as RCM25 to RCM29), and the power electronics problems (denoted as RCM30 to RCM35), respectively. All these benchmark test problems are scalable to any number of objectives while we consider $m\in\{2,3,5,10\}$ for C-DTLZ, DC-DTLZ problems and $m\in\{2,3,4,5\}$ for RWCMOPs in our experiments. The mathematical definitions of these benchmark test problems along with settings of the number of variables and the number of constraints can be found in the supplemental document of this paper\footnote{The supplemental document can be downloaded from \href{https://tinyurl.com/23dtdne8}{here}.}.

\subsection{Peer Algorithms and Parameter Settings}
\label{sec:peer_algorithms}

In our empirical study, we choose to investigate the performance of four widely studied EMO algorithms for CMOPs, including \texttt{C-NSGA-II}~\cite{DebAPM02}, \texttt{C-NSGA-III}~\cite{JainD14}, \texttt{C-MOEA/D}~\cite{JainD14}, and \texttt{C-TAEA}~\cite{LiCFY19}. To address our overarching research question stated at the end of~\pref{sec:introduction}, we design a variant for each of these peer algorithms (dubbed $v$\texttt{C-NSGA-II}, $v$\texttt{C-NSGA-III}, $v$\texttt{C-MOEA/D}, and $v$\texttt{C-TAEA}, respectively) by replacing the CV with a crisp value. Specifically, if a solution $\mathbf{x}$ is feasible, we have $CV(\mathbf{x})=1$; otherwise we set $CV(\mathbf{x})=-1$. The settings of population size and the maximum number of function evaluations are detailed in the supplemental document of this paper~\cite{LiZZL09,LiZLZL09,CaoWKL11,LiKWCR12,LiKCLZS12,LiKWTM13,LiK14,CaoKWL14,LiFKZ14,LiZKLW14,WuKZLWL15,LiKZD15,LiKD15,LiDZ15,LiDZZ17,WuKJLZ17,WuLKZZ17,LiDY18,ChenLY18,ChenLBY18,WuLKZZ19,LiCSY19,Li19,GaoNL19,LiXT19,ZouJYZZL19,LiuLC20,LiXCT20,LiLDMY20,WuLKZ20,BillingsleyLMMG19,LiX0WT20,WangYLK21,BillingsleyLMMG21,YangHL21,ChenLTL22,LiLLM21,LaiL021}.

%We compare the performance of four state-of-the-art algorithms in the situation with/without the assistance of CV values, including C-NSGA-II, C-NSGA-III, C-MOEA/D, and C-TAEA. We do not intend to delineate their working mechanisms here while interested readers are referred to their original papers for details. All algorithms are implemented 31 times independently on each problem to calculate the statistical data for the assessment of performance, the parameter settings of each run are listed as follows:

%(1) Number of Maximum function evaluations (\begin{math} {MFEs} \end{math}) for C-DTLZ and DC-DTLZ test problems: For $ M=2,3, 5, 10$, the population-size of each algorithms is set to 91, 91, 210, and 275, respectively. Moreover, the \begin{math} {MFEs} \end{math} for each problem set as the Table 2 of Appendix.

%(2) Number of \begin{math} {MFEs} \end{math} for RWCMOPs test problems: For $M=2, 3, 4, 5$, the population-size of each algorithms is set to 80, 105, 143, and 212, respectively. For 
%\begin{math} 
%{D\le{10 }}
%\end{math}, 
%and 
%\begin{math}
%{D > 10}
%\end{math}, 
%the \begin{math} {MFEs} \end{math} is fixed at 2500 and 10000. Therefore, the \begin{math} {MFEs} \end{math} for each problem set as the Table 1 of Appendix.

\subsection{Performance Metrics and Statistical Tests}
\label{sec:metrics}

This paper applies the widely used inverted generational distance (IGD)~\cite{BosmanT03}, IGD$^+$~\cite{IshibuchiMTN15}, and hypervolume (HV)~\cite{ZitzlerT99} as the performance metrics to evaluate the performance of different peer algorithms. In our empirical study, each experiment is independently repeated $31$ times with a different random seed. To have a statistical interpretation of the significance of comparison results, we use the following two statistical measures in our empirical study.
\begin{itemize}
	\item\underline{Wilcoxon signed-rank test}~\cite{Wilcoxon1945IndividualCB}: This is a non-parametric statistical test that makes no assumption about the underlying distribution of the data and has been recommended in many empirical studies in the EA community~\cite{DerracGMH11}. In particular, the significance level is set to $p=0.05$ in our experiments.
%    \item\underline{Scott-Knott test}~\cite{MittasA13}: Instead of merely comparing the raw HV values, we apply the Scott-Knott test to rank the performance of different peer techniques over $31$ runs on each test problem. In a nutshell, the Scott-Knott test uses a statistical test and effect size to divide the performance of peer algorithms into several clusters. In particular, the performance of peer algorithms within the same cluster are statistically equivalent. Note that the clustering process terminates until no split can be made. Finally, each cluster can be assigned a rank according to the mean HV and IGD+ values achieved by the peer algorithms within the cluster. In particular, the smaller the rank is, the better performance of the algorithm achieves.
    \item\underline{$A_{12}$ effect size}~\cite{VarghaD00}: To ensure the resulted differences are not generated from a trivial effect, we apply $A_{12}$ as the effect size measure to evaluate the probability that one algorithm is better than another. Specifically, given a pair of peer algorithms, $A_{12}=0.5$ means they are \textit{equal}. $A_{12}>0.5$ denotes that one is better for more than 50\% of the times. $0.56\leq A_{12}<0.64$ indicates a \textit{small} effect size while $0.64 \leq A_{12} < 0.71$ and $A_{12} \geq 0.71$ mean a \textit{medium} and a \textit{large} effect size, respectively. 
\end{itemize}

\section{Experimental Results}
\label{sec:results}

The PFs and the feasible regions of the synthetic problems are relatively simple whereas those of RWCMOPs are complex. In this section, we plan to separate the discussion on the synthetic problems (i.e., C-DTLZ and DC-DTLZ) from the RWCMOPs in view of their distinctive characteristics. 

\subsection{Performance Analysis on Synthetic Benchmark Test Problems}
\label{sec:result_synthetic}

\begin{table*}[t!]
    \caption{Summary of the Wilcoxon signed-rank test results of four selected EMO algorithms against their corresponding variants on IGD, IGD$^+$, and HV.}
    \label{tab:overall_result}
    \centering
	\resizebox{.9\textwidth}{!}{ 
    \begin{tabular}{c|c|c|c|c|c}
        \hline
        \multirow{2}[0]{*}{Problems} & \multirow{2}[0]{*}{Metrics} & \texttt{C-NSGA-II} & \texttt{C-NSGA-III} & \texttt{C-MOEA/D} & \texttt{C-TAEA} \\ 
        ~ &  ~  & $+/-/=$ & $+/-/=$ & $+/-/=$ & $+/-/=$ \\ \hline
        \multicolumn{1}{c|}{\multirow{3}[0]{*}{C-DTLZ and DC-DTLZ}} & IGD   & 1/7/32 & 1/13/26 & 3/1/36 & 2/0/38 \\
%        C-DTLZs$\&$DC-DTLZs & IGD & 1/7/32 & 1/13/26 & 3/1/36 & 2/0/38 \\ 
        & IGD$^+$ & 1/7/32 & 0/15/25 & 3/7/36 & 1/0/39 \\ 
        & HV & 0/8/32 & 0/15/25 & 4/1/35 & 0/0/40 \\\hline\hline
        RWCMOPs & HV & 0/8/27 & 0/11/24 & 4/7/24 & 1/4/30 \\ 
        \hline
    \end{tabular}
    }
    \begin{tablenotes}
    \item[1] $+$, $-$, and $=$ denote the performance of the selected algorithm is significantly better, worse, and equivalent to the corresponding variant, respectively.
    \end{tablenotes} 
\end{table*}

\begin{figure}[t!]
    \centering
    \includegraphics[width=.9\textwidth]{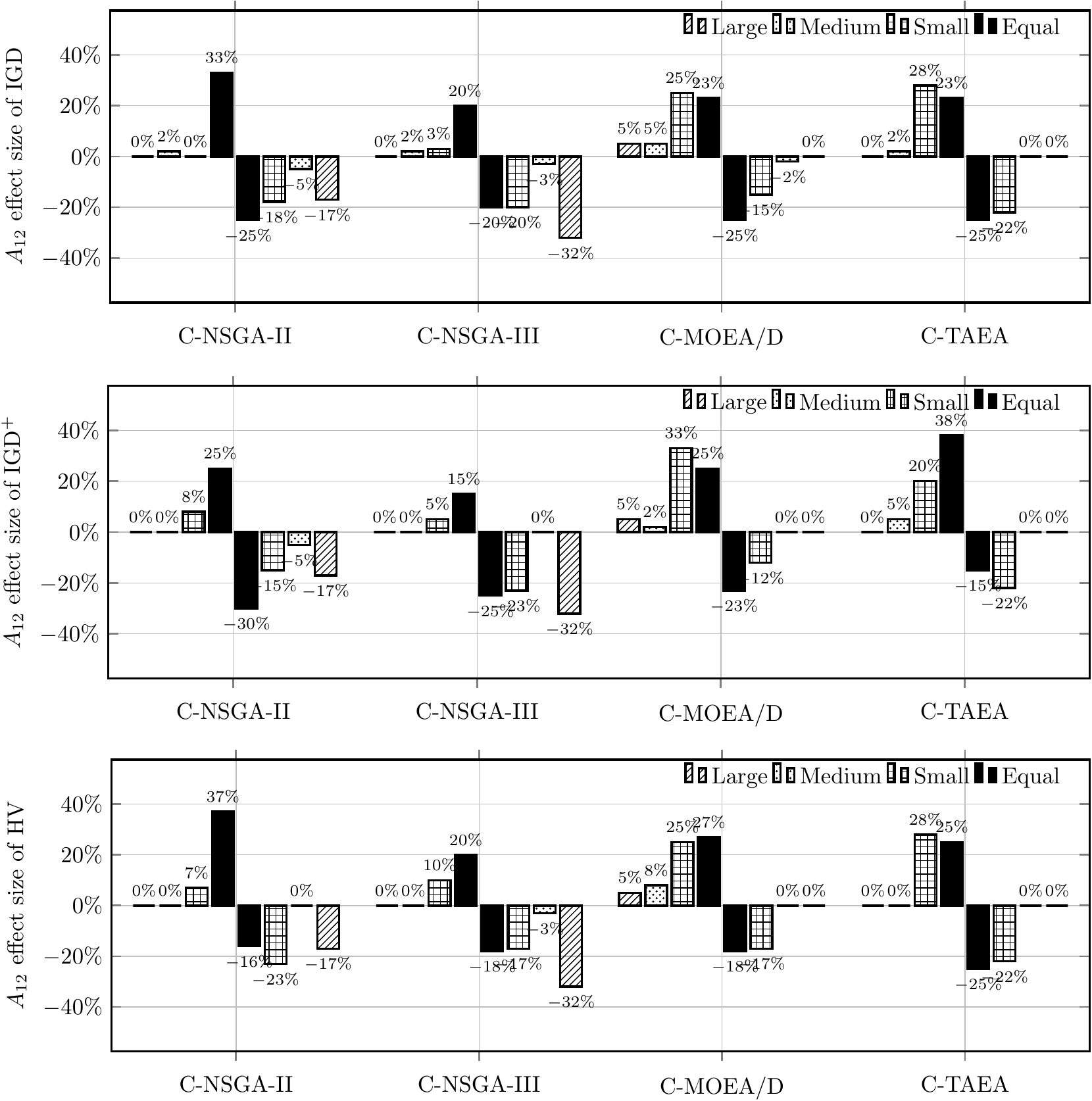}
    \caption{Percentage of the large, medium, small, and equal A12 effect size of metrics for C-DTLZ and DC-DTLZ problems. $+$ means that the variant that replaces the CV with a crisp value can obtain a better result; $-$ means the opposite case.}
    \label{fig:A12DTLZ}
\end{figure}

\begin{figure}[t!]
    \centering
    \includegraphics[width=\textwidth]{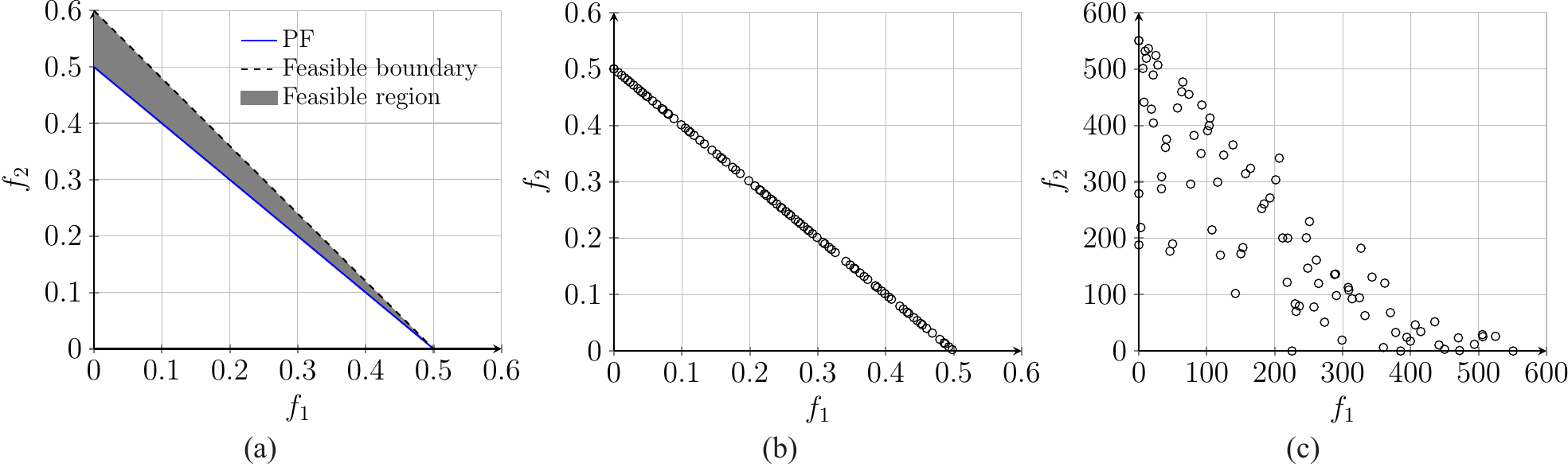}
    \caption{(a) The illustration of the feasible region of C1-DTLZ1; (b) and (c) are the scatter plots of the non-dominated solutions (with the median IGD value) obtained by \texttt{C-NSGA-II} and $v$\texttt{C-NSGA-II}, respectively.}
    \label{fig:C1DTLZ1}
\end{figure}

\begin{figure}[t!]
    \centering
    \includegraphics[width=\textwidth]{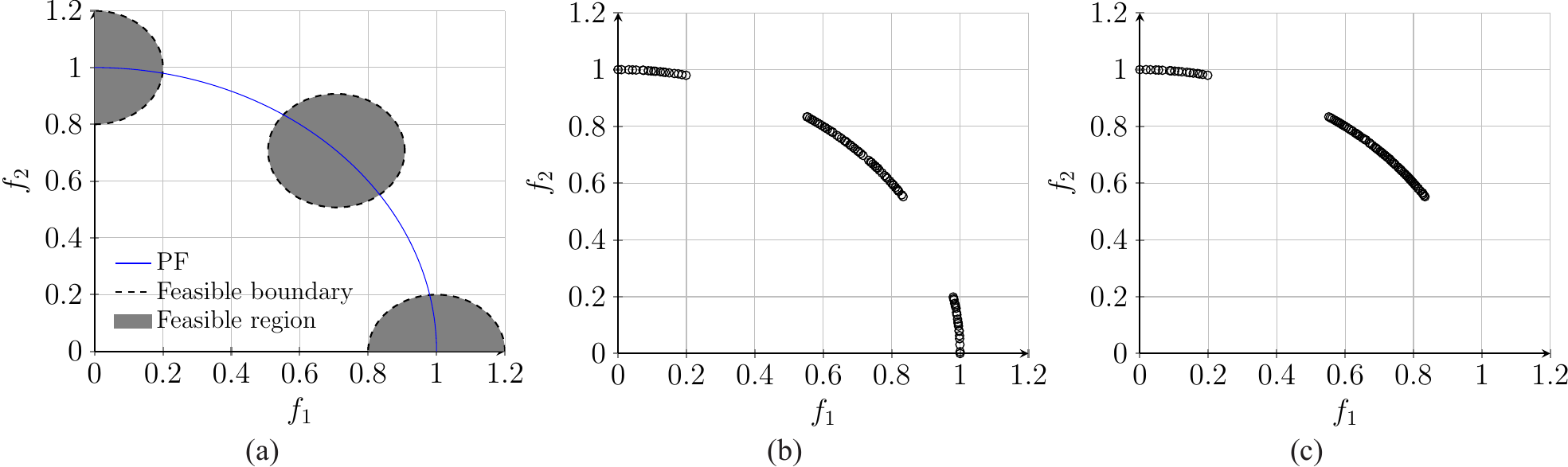}
    \caption{(a) The illustration of the feasible region of C2-DTLZ2; (b) and (c) are the scatter plots of the non-dominated solutions (with the median IGD value) obtained by \texttt{C-NSGA-II} and $v$\texttt{C-NSGA-II}, respectively.}
    \label{fig:C2DTLZ2}
\end{figure}

Due to page limit, we leave the complete comparison results of IGD, IGD$^+$, and HV in Table 3 to Table 8 of supplemental document. Instead, we summarize the Wilcoxon signed-rank test results in the~\pref{tab:overall_result}. From this table, it is clear to see that most comparison results (at least 62.5\% and it even goes to 100\% for the HV comparisons between C-TAEA and $v$C-TAEA) do not have any statistical significance. In other words, replacing the CV with a crisp value does not significantly influence the performance on C-DTLZ and DC-DTLZ problems. In addition to the pairwise comparisons, we apply the $A_{12}$ effect size to have a better understanding of the performance difference between the selected EMO algorithm and its corresponding variant. From the collected comparison results ($50\times2=100$ in total) shown in~\pref{fig:A12DTLZ}, we can see that most of the comparison results are classified as $equal$ (ranging from 38\% to 58\%). As reflected in~\pref{tab:overall_result}, it is surprising to see that the corresponding variants (i.e., without using the CV to guide the evolutionary search process) have achieved better performance in many cases. In particular, up to 32\% comparison  results are classified to be $large$. In the following paragraphs, we plan to analyse some remarkable findings collected from the results.
\begin{itemize}
   \item Let us first look into the performance of \texttt{C-NSGA-II} and \texttt{C-NSGA-III} w.r.t. their variants \texttt{$v$C-NSGA-II} and \texttt{$v$C-NSGA-III}. As shown in~\pref{tab:overall_result}, the performance of \texttt{C-NSGA-II} and \texttt{C-NSGA-III} have been deteriorated (ranging from 17.5\% to 37.5\%) when replacing the CV with a crisp value in their corresponding CHTs, especially on C1-DTLZ1 and C2-DTLZ2.
   \begin{itemize}
      \item As the illustrative example shown in~\pref{fig:C1DTLZ1}, the feasible region of C1-DTLZ1 is a narrow wedge arrow right above the PF. Without the guidance of the CV, both \texttt{C-NSGA-II} and \texttt{C-NSGA-III} become struggling in the large infeasible region. In particular, there is no sufficient selection pressure to guide the population to move forward.
      \item C2-DTLZ2 has several disparately distributed feasible regions as the illustrative example shown in \pref{fig:C2DTLZ2}. Since the CHTs of both \texttt{C-NSGA-II} and \texttt{C-NSGA-III} do not have a dedicated diversity maintenance mechanism, the evolutionary population can be guided towards some, but not all, local feasible region(s) as the examples shown in~\pref{fig:C2DTLZ2}(c).
      \item As for the other test problems, we find that the replacement of CV with a crisp value dose not make a significant impact to the performance of both \texttt{C-NSGA-II} and \texttt{C-NSGA-III}. This can be explained as a large feasible region that makes the Pareto dominance alone can provide sufficient selection pressure towards the PF.
   \end{itemize}
   \item It is interesting to note that \texttt{C-MOEA/D} uses the same CHT as \texttt{C-NSGA-II} and \texttt{C-NSGA-III}, but its performance does not deteriorate significantly when replacing the CV with a crisp value as shown in~\pref{tab:overall_result}. This can be understood as the baseline \texttt{MOEA/D} that provides a better mechanism to preserve the population diversity during the environmental selection. Thus, the evolutionary population can overcome the infeasible regions towards the PF.
   \item As for \texttt{C-TAEA}, it is surprising to note that the consideration of the CV does not pose any impact to its performance as evidenced in~\pref{tab:overall_result} (nearly all comparison results have no statistical significance). This can be explained as the use of the diversity-oriented archive in \texttt{C-TAEA} which does not consider the CV but just relies on the Pareto dominance along to drive the evolutionary population.
\end{itemize}

\subsection{Performance Analysis on Real-World Benchmark Test Problems}
\label{sec:result_rwcmops}

\begin{figure}[t!]
    \centering
    \includegraphics[width=.9\textwidth]{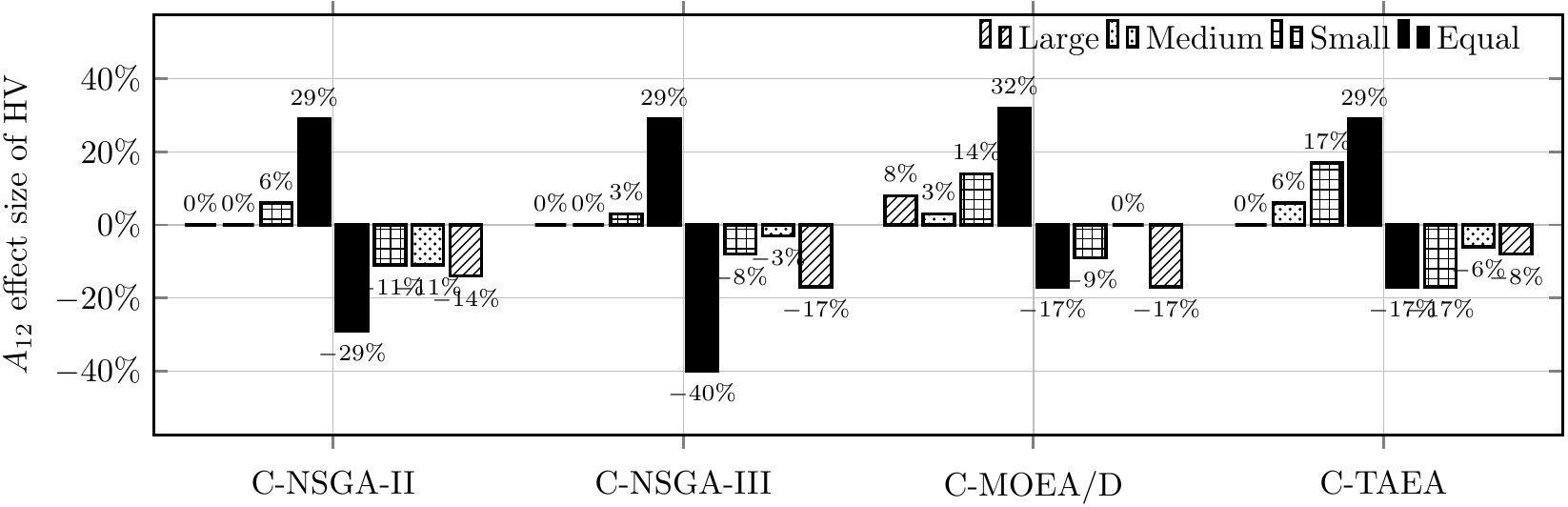}
    \caption{Percentage of the large, medium, small, and equal $A_{12}$ effect size of metrics for RWCMOPs. $+$ means that the variant that replaces the CV with a crisp value can obtain a better result; $-$ means the opposite case.}
    \label{fig:a12_rwcmops}
\end{figure}

Since the PFs of the RWCMOPs are unknown a priori, we only apply the HV as the performance metric in this study. As in~\pref{sec:result_synthetic}, the complete comparison results of HV are given in Tables 9 and 10 of the supplemental document while the Wilcoxon signed-rank test results are summarized in~\pref{tab:overall_result}. From these results, we can see that most of the comparisons (around 68.5\% to 85.7\%) do not have statistical significance. In other words, there is a marginal difference when replacing the CV with a crisp value. To have a better understanding of the difference, we again apply the $A_{12}$ effect size to complement the results of the Wilcoxon signed-rank test. From the bar charts shown in~\pref{fig:a12_rwcmops}, it is clear to see that most comparison results (ranging from 46\% to 69\%) are classified to be equal while only up to 14\% comparison results are classified to have a large difference. In the following paragraphs, we plan to elaborate some selected results on problems with an equal and a large effect size, respectively.

\begin{figure}[t!]
    \centering
    \includegraphics[width=\textwidth]{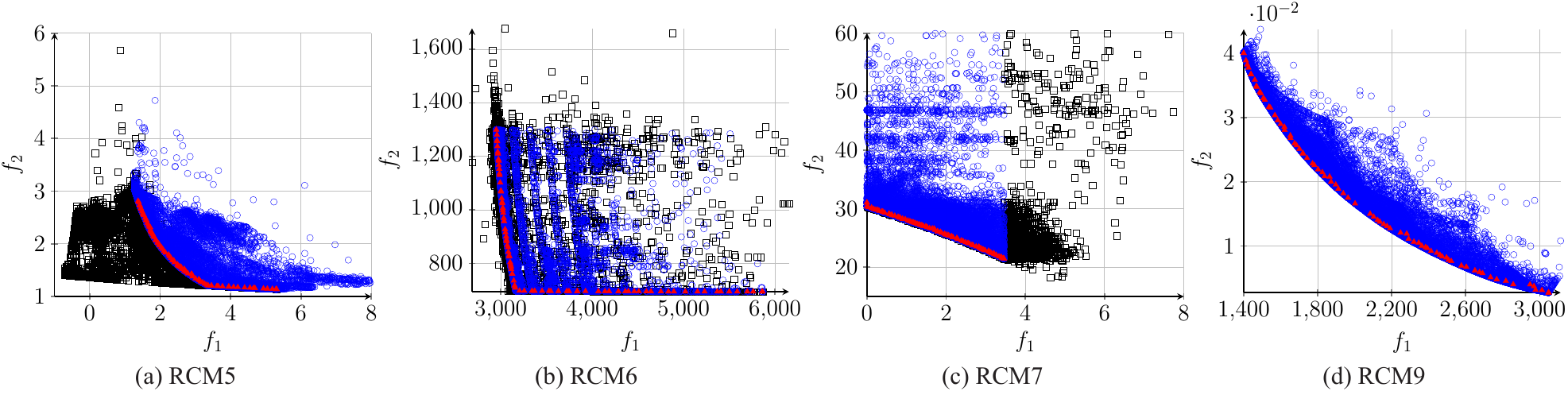}
    \caption{Distribution of feasible solutions (denoted as the \textcolor{blue}{blue circle}), infeasible solutions (denoted as the black square), and non-dominated solutions (denoted as \textcolor{red}{red triangle}) obtained by \texttt{C-NSGA-II} on RCM5, RCM6, RCM7 and RCM9.}
    \label{fig:equal_rwcmops}
\end{figure}

As for the RWCMOPs whose $A_{12}$ effect size comparison results are classified as equal, we consider the following four test problems in our analysis.
\begin{itemize}
    \item Let us start from the RCM5 problem. As shown in Fig~\ref{fig:equal_rwcmops}(a), the feasible and infeasible regions have almost the same size while the PF is located in the intersection between them. In this case, it is natural that the environmental selection can provide necessary selection pressure without using the CV.

    \item As for the RCM6 problem shown in~\pref{fig:equal_rwcmops}(b), the feasible and infeasible regions are intertwined with each other. Therefore, the infeasible region does not really provide an obstacle to the evolutionary population. Accordingly, the CV plays a marginal role for constraint handling.

    \item Similar to the RCM5 problem, the RCM7 problem has a large and opening feasible region as shown in~\pref{fig:equal_rwcmops}(c). In addition, the feasible and infeasible regions are hardly overlapped with each. In this case, the evolutionary population can have a large chance to explore in the feasible region without any interference from the infeasible solutions.

    \item At the end, as shown in~\pref{fig:equal_rwcmops}(d), it is hardly to treat the RCM9 problem as a CMOP since the feasible region is overtaking the infeasible region. In other words, the feasible region is too large to find an infeasible solution. Accordingly, it is not difficult to understand that the CV becomes useless.
\end{itemize}

\begin{figure}[t!]
    \centering
    \includegraphics[width=\textwidth]{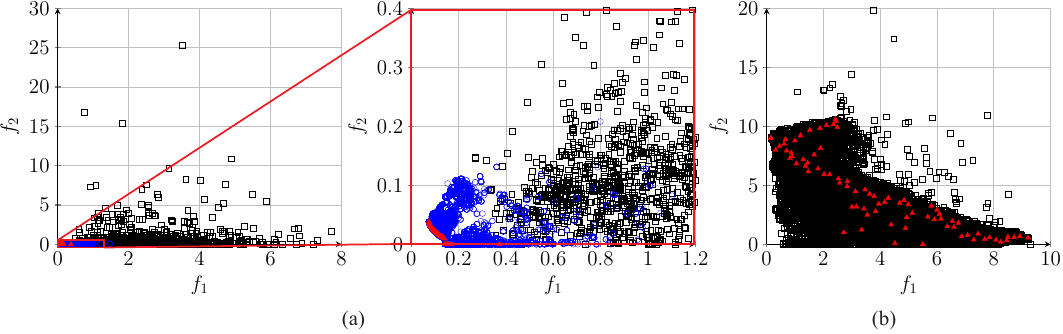}
    \caption{(a) Distribution of feasible solutions (denoted as the \textcolor{blue}{blue circle}), infeasible solutions (denoted as the black square), and non-dominated solutions (denoted as \textcolor{red}{red triangle}) obtained by \texttt{C-NSGA-II} on RCM30. (b) Distribution of infeasible solutions (denoted as the black square) and non-dominated solutions (denoted as \textcolor{red}{red triangle}) obtained by $v$\texttt{C-NSGA-II} on RCM30.}
    \label{fig:large_rcm30}
\end{figure}

\begin{figure}[t!]
    \centering
    \includegraphics[width=\textwidth]{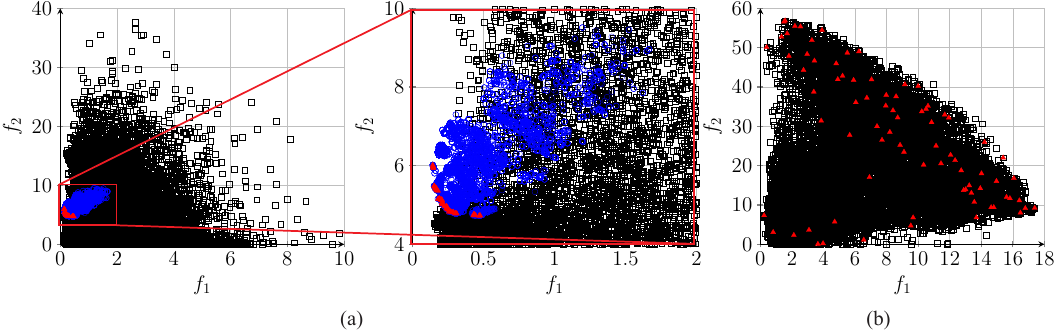}
    \caption{(a) Distribution of feasible solutions (denoted as the \textcolor{blue}{blue circle}), infeasible solutions (denoted as the black square), and non-dominated solutions (denoted as \textcolor{red}{red triangle}) obtained by \texttt{C-NSGA-II} on RCM35. (b) Distribution of infeasible solutions (denoted as the black square) and non-dominated solutions (denoted as \textcolor{red}{red triangle}) obtained by $v$\texttt{C-NSGA-II} on RCM35.}
    \label{fig:large_rcm35}
\end{figure}

As for the other RWCMOPs, of which the comparison results are classified to be large according to the $A_{12}$ effect size, we pick up two remarkable cases and make some analysis as follows. 
\begin{itemize}
    \item Let us first consider the RCM30 problem. As shown in~\pref{fig:large_rcm30}(a), the feasible region of the RCM30 problem is very narrow and is squeezed towards the PF. Therefore, it is not difficult to understand that the evolutionary population can hardly be navigated without the guidance of the CV. As shown in~\pref{fig:large_rcm30}(b), the solutions obtained by $v$\texttt{C-NSGA-II} are far away from the PF.

    \item As shown in~\pref{fig:large_rcm35}(a), comparing to the RCM30 problem, the size of the feasible region of the RCM35 problem is much wider. However, it is still largely surrounded by the infeasible region. In this case, as shown in~\pref{fig:large_rcm35}(b), without the guidance of the CV, the evolutionary population can not only have sufficient selection pressure to move forward, but also can be mislead to the infeasible region that dominates the feasible region.
\end{itemize}


%!TeX root=main.tex

\section{Conclusion}
\label{sec:conclusion}

Most, if not all, existing CHT in EMO implicitly assume that the formulation of the constraint function(s) of a CMOP is well defined a priori. Therefore, the CV has been widely used as the building block for designing CHTs to provide an extra selection pressure in the environmental selection. However, this assumption is arguably viable for real-world optimization scenarios of which the problems are treated as a black box. In this case, it is hardly to derive the CV in practice. Bearing this consideration in mind, this paper empirically investigate the impact of replacing the CV with a crisp value in the CHTs of four prevalent EMO algorithms for CMOPs. From our empirical results on both synthetic and real-world benchmark test problems, it is surprising to see that the performance is not significantly deteriorated when the CV is not used to guide the evolutionary population. One of the potential reasons is that the feasible is large enough to attract the evolutionary population thus leading to a marginal obstacle for an EMO algorithm to overcome the infeasible region. This directly comes up to the requirement of new benchmark test problems with more challenging infeasible regions. In addition, this also inspires new research opportunity to develop new CHT(s) to handle the CMOP with unknown constraint in the near future.

\section*{Acknowledgment}
This work was supported by UKRI Future Leaders Fellowship (MR/S017062/1), EPSRC (2404317), NSFC (62076056), Royal Society (IES/R2/212077) and Amazon Research Award.

\bibliographystyle{IEEEtran}
\bibliography{IEEEabrv,cmo}

\end{document}

\title{Supplementary Document of “Do We Really Need to Use Constraint Violation in Constrained Evolutionary Multi-Objective Optimization?”\thanks{K. Li was supported by UKRI Future Leaders Fellowship (Grant No. MR/S017062/1) and Royal Society (Grant No. IEC/NSFC/170243).}}

\author{Shuang Li\inst{1} \and
Ke Li\inst{2}\orcidID{0000-0001-7200-4244}
}
\author{Shuang Li\inst{1} \orcidID{0000-0001-6261-177X}\and
Ke Li\inst{2}\orcidID{0000-0001-7200-4244}\and
Wei Li\inst{1}
}

\authorrunning{L. Shuang et al.}
\institute{Control and Simulation Center, Harbin Institute of Technology, Harbin, China \and
Department of Computer Science, University of Exeter, Exeter, EX4 5DS, UK\\
\email{frank@hit.edu.cn}}

\maketitle   
\section{Experimental Settings}
\subsection{Benchmark Test Problems}
In our empirical study, we pick up $45$ benchmark test problems widely studied in the literature to constitute our benchmark suite. More specifically, it consists of C1-DTLZ1, C1-DTLZ3, C2-DTLZ2 and C3-DTLZ4 from the C-DTLZ benchmark suite~\cite{JainD14}; DC1-DTLZ1, DC1-DTLZ3, DC2-DTLZ1, DC2-DTLZ3, DC3-DTLZ1, DC3-DTLZ3 chosen from the DC-DTLZ benchmark suite~\cite{LiCFY19}; and other $35$ problems picked up from the real-world constrained multi-objective problems (RWCMOPs) benchmark suite~\cite{KumarWALMSD21}. In particular, the RWCMOPs are derived from the mechanical design problems (denoted as RCM1 to RCM21), the chemical engineering problems (denoted as RCM22 to RCM24), the process design and synthesis problems (denoted as RCM25 to RCM29), and the power electronics problems (denoted as RCM30 to RCM35), respectively. All these benchmark test problems are scalable to any number of objectives while we consider $m\in\{2,3,5,10\}$ for C-DTLZ, DC-DTLZ problems and $m\in\{2,3,4,5\}$ for RWCMOPs in our experiments. The mathematical definitions of these benchmark test problems can be found in their original papers. The basic details of RWCMOPs such as the number of objective functions, number of decision variables, number of equality constraints (${nh}$), and inequality constraints (${ng}$) are shown in Table~\ref{tab:RWCMOP}.

\begin{figure}%[htbp]
  \centering
	\includegraphics[width=\textwidth]{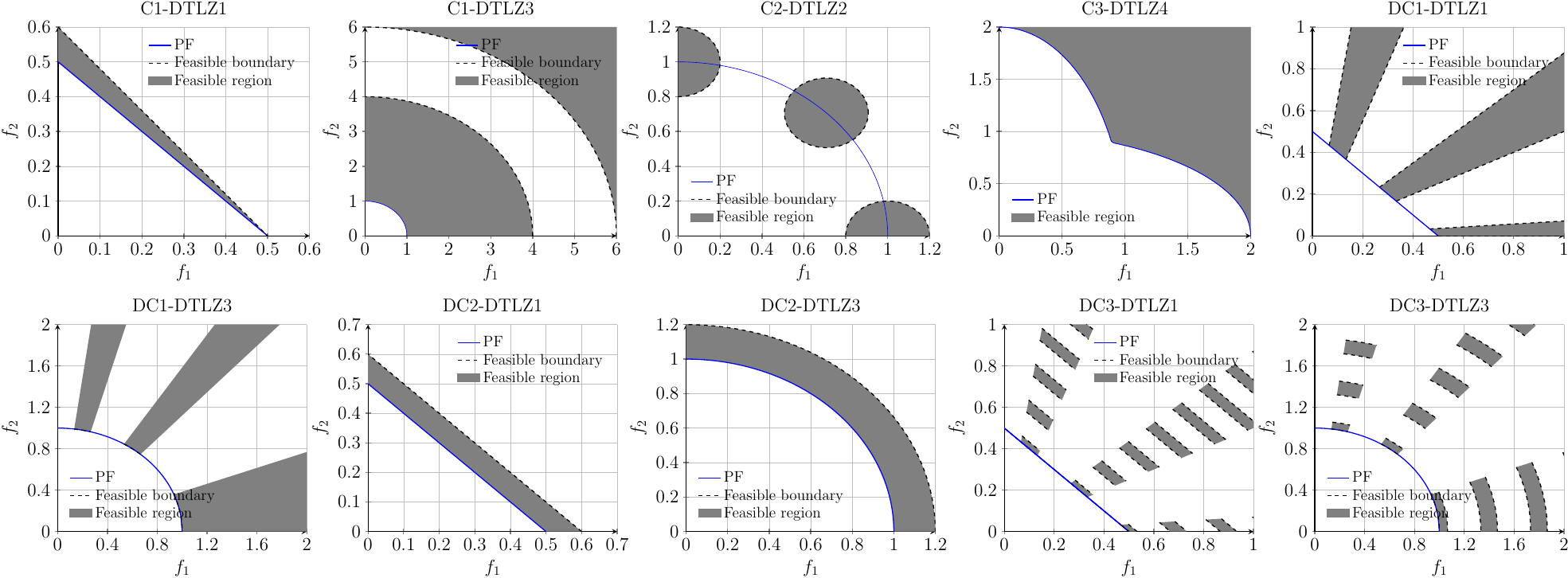}
  \caption{The illustrations of the feasible region of C-DTLZ and DC-DTLZ benchmark test problems}
  \label{RCM-C-TAEA}
\end{figure}

\begin{table}[h]
  \scriptsize
  \centering
  \caption{Details of RWCMOPs test instances.}
  \label{tab:RWCMOP}
    \begin{tabular}{llllllllllll}
    \toprule
     Problem & M & D & ng & nh & MFE & Problem & M & D & ng & nh & MFE\\
    \midrule
        RWMOP1 & 2 & 4 & 2 & 2 & 20000 & RWMOP19 & 3 & 10 & 10 & 0 & 26250 \\
        RWMOP2 & 2 & 5 & 5 & 0 & 20000 & RWMOP20 & 2 & 4 & 7 & 0 & 20000 \\ 
        RWMOP3 & 2 & 3 & 3 & 0 & 20000 & RWMOP21 & 2 & 6 & 4 & 0 & 20000 \\ 
        RWMOP4 & 2 & 4 & 4 & 0 & 20000 & RWMOP22 & 2 & 9 & 2 & 4 & 20000 \\ 
        RWMOP5 & 2 & 4 & 4 & 0 & 20000 & RWMOP23 & 2 & 6 & 1 & 4 & 20000 \\ 
        RWMOP6 & 2 & 7 & 11 & 0 & 20000 & RWMOP24 & 3 & 9 & 0 & 6 & 26250 \\ 
        RWMOP7 & 2 & 4 & 1 & 0 & 20000 & RWMOP25 & 2 & 2 & 2 & 0 & 20000 \\ 
        RWMOP8 & 3 & 7 & 9 & 0 & 26250 & RWMOP26 & 2 & 3 & 1 & 1 & 20000 \\ 
        RWMOP9 & 2 & 4 & 0 & 0 & 20000 & RWMOP27 & 2 & 3 & 3 & 0 & 20000 \\ 
        RWMOP10 & 2 & 2 & 2 & 0 & 20000 & RWMOP28 & 2 & 7 & 4 & 4 & 20000 \\ 
        RWMOP11 & 5 & 3 & 7 & 0 & 53000 & RWMOP29 & 2 & 7 & 9 & 0 & 20000 \\ 
        RWMOP12 & 2 & 4 & 1 & 0 & 20000 & RWMOP30 & 2 & 25 & 24 & 0 & 80000 \\ 
        RWMOP13 & 3 & 7 & 11 & 0 & 20000 & RWMOP31 & 2 & 25 & 24 & 0 & 80000 \\ 
        RWMOP14 & 2 & 5 & 8 & 0 & 26250 & RWMOP32 & 2 & 25 & 24 & 0 & 80000 \\ 
        RWMOP15 & 2 & 3 & 8 & 0 & 20000 & RWMOP33 & 2 & 30 & 29 & 0 & 80000 \\ 
        RWMOP16 & 2 & 2 & 2 & 0 & 20000 & RWMOP34 & 2 & 30 & 29 & 0 & 80000 \\ 
        RWMOP17 & 3 & 6 & 9 & 0 & 26250 & RWMOP35 & 2 & 30 & 29 & 0 & 80000 \\ 
        RWMOP18 & 2 & 3 & 3 & 0 & 20000 & ~ & ~ & ~ & ~ & ~ & ~ \\ 
  \bottomrule
\end{tabular}%
\end{table}%
\vspace{-1em}

\subsection{Peer Algorithms and Parameter Settings}
In our empirical study, we choose to investigate the performance of four widely studied EMO algorithms for CMOPs, including \texttt{C-NSGA-II}~\cite{DebAPM02}, \texttt{C-NSGA-III}~\cite{JainD14}, \texttt{C-MOEA/D}~\cite{JainD14}, and \texttt{C-TAEA}~\cite{LiCFY19}. To address our overarching research question, we design a variant for each of these peer algorithms (dubbed $v$\texttt{C-NSGA-II}, $v$\texttt{C-NSGA-III}, $v$\texttt{C-MOEA/D}, and $v$\texttt{C-TAEA}, respectively) by replacing the CV with a crisp value. Specifically, if a solution $\mathbf{x}$ is feasible, we have $CV(\mathbf{x})=1$; otherwise we set $CV(\mathbf{x})=-1$. The settings of population size and the maximum number of function evaluations are listed as follows:

(1) Number of Maximum function evaluations (\begin{math} {MFEs} \end{math}) for C-DTLZ and DC-DTLZ test problems: For $ M=2,3, 5, 10$, the population-size of each algorithms is set to 91, 91, 210, and 275, respectively. Moreover, the \begin{math} {MFEs} \end{math} for each problem set as the Table~\ref{tab:MFEs1}.

\begin{table}[h]
  \scriptsize
  \centering
  \caption{Number of MFEs of Different C-DTLZ and DC-DTLZ Test Problems.}
  \label{tab:MFEs1}
  \begin{tabular}{lllll}
    \toprule
    Problem & M=2 & M=3 & M=5 & M=10 \\
    \midrule
    C1\_DTLZ1 & 500*N & 500*N & 600*N & 1000*N \\
    C1\_DTLZ3 & 1000*N & 1000*N & 1500*N & 3000*N \\
    C2\_DTLZ2 & 250*N & 250*N & 350*N & 750*N \\
    C3\_DTLZ4 & 750*N & 750*N & 1250*N & 3000*N \\
    DC1\_DTLZ1 & 600*N & 800*N & 600*N & 800*N \\
    DC1\_DTLZ3 & 600*N & 1000*N & 1000*N & 1200*N \\
    DC2\_DTLZ1 & 700*N & 700*N & 500*N & 600*N \\
    DC2\_DTLZ3 & 900*N & 1100*N & 1400*N & 1500*N \\
    DC3\_DTLZ1 & 700*N & 700*N & 600*N & 600*N \\
    DC3\_DTLZ3 & 1000*N & 1300*N & 1400*N & 1800*N \\
    %\O & 1 in 1,000& For Swedish names\\
    %$\pi$ & 1 in 5& Common in math\\
    %\$ & 4 in 5 & Used in business\\
    %$\Psi^2_1$ & 1 in 40,000& Unexplained usage\\
  \bottomrule
\end{tabular}
\end{table}

(2) Number of \begin{math} {MFEs} \end{math} for RWCMOPs test problems: For $M=2, 3, 4, 5$, the population-size of each algorithms is set to 80, 105, 143, and 212, respectively. The \begin{math} {MFEs} \end{math} for each problem set as the Table~\ref{tab:RWCMOP}.

\subsection{Performance Metrics and Statistical Tests}
\label{sec:metrics}

This paper applies the widely used inverted generational distance (IGD)~\cite{BosmanT03}, IGD$^+$~\cite{IshibuchiMTN15}, and hypervolume (HV)~\cite{ZitzlerT99} as the performance metrics to evaluate the performance of different peer algorithms. In our empirical study, each experiment is independently repeated $31$ times with a different random seed. To have a statistical interpretation of the significance of comparison results, we use the following three statistical measures in our empirical study.
\begin{itemize}
	\item\underline{Wilcoxon signed-rank test}~\cite{Wilcoxon1945IndividualCB}: This is a non-parametric statistical test that makes no assumption about the underlying distribution of the data and has been recommended in many empirical studies in the EA community~\cite{DerracGMH11}. In particular, the significance level is set to $p=0.05$ in our experiments.
    \item\underline{$A_{12}$ effect size}~\cite{VarghaD00}: To ensure the resulted differences are not generated from a trivial effect, we apply $A_{12}$ as the effect size measure to evaluate the probability that one algorithm is better than another. Specifically, given a pair of peer algorithms, $A_{12}=0.5$ means they are \textit{equal}. $A_{12}>0.5$ denotes that one is better for more than 50\% of the times. $0.56\leq A_{12}<0.64$ indicates a \textit{small} effect size while $0.64 \leq A_{12} < 0.71$ and $A_{12} \geq 0.71$ mean a \textit{medium} and a \textit{large} effect size, respectively. 
\end{itemize}

\section{Experimental results}
In this section, we plan to separate the results on the synthetic problems (i.e., C-DTLZ and DC-DTLZ) from the RWCMOPs in view of their distinctive characteristics. In a nutshell, the PFs and the feasible regions of the synthetic problems are relatively simple whereas those of RWCMOPs are complex.

\subsection{Results on Synthetic Benchmark Test Problems}
Here, we report the Wilcoxon signed-rank test results in the Table~\ref{tab:DTLZIGDa} to Table~\ref{tab:DTLZHVb}. From these tables, it is clear to see that most comparison results (at least 62.5\% while it goes to 100\% for the HV comparisons between C-TAEA and $v$C-TAEA) do not have any statistical significance. In other words, replacing the CV with a crisp value does not significantly influence the performance on C-DTLZ and DC-DTLZ problems. The final solution set of these synthetic benchmark test problems (with median IGD metric) also support above conclusions, the scatter plots are shown in Figure~\ref{DTLZ-C-NSGA-II}, \ref{DTLZ-C-NSGA-III}, \ref{DTLZ-C-MOEA/D}, \ref{DTLZ-C-TAEA}.

   \begin{table*}%[htbp]
   \begin{footnotesize}
   \begin{center}
  \caption{Comparison results on IGD metric (median and IQR) for C-NSGA-II and C-NSGA-III against their corresponding variants on C-DTLZ and DC-DTLZ synthetic problems}
  \label{tab:DTLZIGDa}
  \centering
  \begin{tabular}{lllllll}
    \toprule
    Problem & M     & $v$C-NSGA-II & C-NSGA-II & $v$C-NSGA-III & C-NSGA-III \\
    C1\_DTLZ1 & 2     & 5.0000e+2 (0.00e+0) - & 2.4596e-3 (1.60e-4) & 5.0000e+2 (0.00e+0) - & 1.9983e-3 (1.24e-4) \\
    C1\_DTLZ3 & 2     & 5.5804e-3 (4.99e+0) + & 5.0001e+0 (4.99e+0) & 4.3740e-3 (2.00e-5) = & 4.3803e-3 (3.31e-5) \\
    C2\_DTLZ2 & 2     & 1.5126e-1 (2.08e-1) - & 4.5451e-2 (1.06e-1) & 2.5697e-1 (1.06e-1) - & 4.6782e-2 (1.56e-4) \\
    C3\_DTLZ4 & 2     & 1.0889e-2 (6.46e-4) = & 1.0918e-2 (6.44e-4) & 9.3603e-3 (2.70e-4) + & 9.5734e-3 (6.20e-4) \\
    DC1\_DTLZ1 & 2     & 4.7724e-2 (3.91e-2) = & 6.7223e-2 (3.91e-2) & 4.8013e-2 (1.57e-2) = & 4.7861e-2 (1.99e-2) \\
    DC1\_DTLZ3 & 2     & 1.1794e-1 (2.02e-4) = & 1.1790e-1 (1.73e-4) & 1.1966e-1 (6.53e-4) = & 1.1967e-1 (5.98e-4) \\
    DC2\_DTLZ1 & 2     & 5.0000e+2 (0.00e+0) = & 5.0000e+2 (0.00e+0) & 5.0000e+2 (0.00e+0) - & 1.9646e-1 (1.90e-4) \\
    DC2\_DTLZ3 & 2     & 5.0000e+2 (0.00e+0) = & 5.0000e+2 (0.00e+0) & 5.0000e+2 (0.00e+0) - & 5.5562e-1 (0.00e+0) \\
    DC3\_DTLZ1 & 2     & 2.2565e-1 (1.69e-1) = & 2.2565e-1 (1.73e-1) & 2.2568e-1 (1.73e-1) = & 2.2567e-1 (1.73e-1) \\
    DC3\_DTLZ3 & 2     & 1.9096e+0 (6.63e-1) = & 1.9096e+0 (6.63e-1) & 1.2471e+0 (1.15e+0) = & 1.2471e+0 (2.20e-4) \\
    C1\_DTLZ1 & 3     & 5.0000e+2 (0.00e+0) - & 2.8156e-2 (1.73e-3) & 5.0000e+2 (0.00e+0) - & 2.0311e-2 (2.02e-4) \\
    C1\_DTLZ3 & 3     & 8.0234e+0 (4.74e-3) = & 8.0248e+0 (5.58e+0) & 1.2302e-1 (7.96e+0) = & 8.0070e+0 (7.96e+0) \\
    C2\_DTLZ2 & 3     & 4.6619e-1 (5.66e-1) - & 1.2759e-1 (4.92e-3) & 2.7357e-1 (8.93e-2) - & 1.2174e-1 (1.01e-2) \\
    C3\_DTLZ4 & 3     & 1.3165e-1 (7.37e-3) = & 1.3334e-1 (6.34e-3) & 9.2496e-2 (1.06e-3) = & 9.2605e-2 (6.16e-4) \\
    DC1\_DTLZ1 & 3     & 5.3563e-2 (1.94e-2) = & 7.0010e-2 (1.98e-2) & 5.1665e-2 (2.19e-2) = & 5.1629e-2 (2.17e-2) \\
    DC1\_DTLZ3 & 3     & 1.3467e-1 (4.10e-3) = & 1.3527e-1 (4.94e-3) & 1.2367e-1 (7.10e-4) = & 1.2391e-1 (8.87e-4) \\
    DC2\_DTLZ1 & 3     & 5.0000e+2 (0.00e+0) = & 5.0000e+2 (0.00e+0) & 5.0000e+2 (0.00e+0) - & 1.6428e-1 (5.31e-4) \\
    DC2\_DTLZ3 & 3     & 5.0000e+2 (0.00e+0) = & 5.0000e+2 (0.00e+0) & 5.0000e+2 (0.00e+0) = & NaN (NaN) \\
    DC3\_DTLZ1 & 3     & 1.8863e-1 (8.16e-4) = & 1.8852e-1 (7.84e-2) & 1.9058e-1 (1.12e-1) = & 1.9130e-1 (1.68e-3) \\
    DC3\_DTLZ3 & 3     & 1.2640e+0 (6.62e-1) = & 1.2659e+0 (6.61e-1) & 1.2654e+0 (3.19e-3) = & 1.2646e+0 (3.69e-3) \\
    C1\_DTLZ1 & 5     & 5.0000e+2 (0.00e+0) - & 6.2527e-2 (1.21e-3) & 0.0000e+0 (0.00e+0) - & 9.7462e-1 (5.94e-3) \\
    C1\_DTLZ3 & 5     & 1.1638e+1 (2.75e-2) = & 1.1643e+1 (2.35e-2) & 0.0000e+0 (0.00e+0) = & 0.0000e+0 (6.25e-1) \\
    C2\_DTLZ2 & 5     & 9.4615e-1 (8.18e-4) = & 4.4088e-1 (5.09e-1) & 0.0000e+0 (0.00e+0) - & 1.0907e-1 (3.29e-1) \\
    C3\_DTLZ4 & 5     & 4.2320e-1 (1.64e-2) = & 4.2735e-1 (2.95e-2) & 9.6342e-1 (1.08e-3) = & 9.6343e-1 (1.45e-3) \\
    DC1\_DTLZ1 & 5     & 8.0433e-2 (2.23e-3) = & 8.0450e-2 (1.91e-3) & 9.7500e-1 (4.09e-3) - & 9.7530e-1 (2.79e-4) \\
    DC1\_DTLZ3 & 5     & 2.4620e-1 (1.61e-2) = & 2.4993e-1 (1.21e-2) & 7.7856e-1 (1.46e-3) = & 7.7832e-1 (1.03e-3) \\
    DC2\_DTLZ1 & 5     & 5.0000e+2 (0.00e+0) = & 5.0000e+2 (0.00e+0) & 0.0000e+0 (0.00e+0) - & 8.3132e-1 (3.31e-3) \\
    DC2\_DTLZ3 & 5     & 5.0000e+2 (0.00e+0) = & 5.0000e+2 (0.00e+0) & 0.0000e+0 (0.00e+0) = & 0.0000e+0 (0.00e+0) \\
    DC3\_DTLZ1 & 5     & 1.1097e-1 (6.48e-2) = & 1.7447e-1 (6.62e-2) & 4.7244e-1 (2.25e-1) = & 4.7091e-1 (2.30e-1) \\
    DC3\_DTLZ3 & 5     & 6.9124e-1 (4.74e-1) = & 6.8739e-1 (1.28e-2) & 0.0000e+0 (0.00e+0) = & 0.0000e+0 (0.00e+0) \\
    C1\_DTLZ1 & 10    & 5.0000e+2 (0.00e+0) - & 1.2160e-1 (3.20e-3) & 5.0000e+2 (0.00e+0) - & 1.0888e-1 (5.14e-4) \\
    C1\_DTLZ3 & 10    & 1.0344e+2 (3.83e+1) = & 8.7379e+1 (2.41e+1) & 1.8137e+0 (1.38e+1) = & 1.3123e+0 (1.38e+1) \\
    C2\_DTLZ2 & 10    & 1.1043e+0 (4.14e-4) - & 6.1719e-1 (4.88e-1) & 1.0978e+0 (7.50e-3) = & 1.0973e+0 (4.90e-1) \\
    C3\_DTLZ4 & 10    & 2.1047e+0 (6.61e-2) = & 2.0962e+0 (4.95e-2) & 5.6805e-1 (1.80e-3) = & 5.6785e-1 (1.91e-3) \\
    DC1\_DTLZ1 & 10    & 5.7283e+0 (3.68e+0) = & 4.6996e+0 (2.95e+0) & 1.1923e-1 (2.16e-2) = & 1.1326e-1 (1.91e-2) \\
    DC1\_DTLZ3 & 10    & 5.2782e+1 (5.52e+0) = & 5.3490e+1 (6.10e+0) & 5.5843e-1 (3.38e-2) = & 5.6297e-1 (6.70e-2) \\
    DC2\_DTLZ1 & 10    & 5.0000e+2 (0.00e+0) = & 5.0000e+2 (0.00e+0) & 5.0000e+2 (0.00e+0) - & 1.7704e-1 (7.21e-3) \\
    DC2\_DTLZ3 & 10    & 5.0000e+2 (0.00e+0) = & 5.0000e+2 (0.00e+0) & 9.3023e-1 (0.00e+0) = & 8.6422e-1 (4.83e-2) \\
    DC3\_DTLZ1 & 10    & 1.9275e-1 (1.30e-1) = & 1.9484e-1 (7.42e-2) & 1.9848e-1 (6.17e-2) = & 1.9742e-1 (1.31e-2) \\
    DC3\_DTLZ3 & 10    & 4.7123e+1 (1.85e+1) = & 4.6957e+1 (2.01e+1) & 1.4419e+0 (6.93e-1) = & 1.4440e+0 (1.49e-1) \\
    +/-/= &   & 1/7/32 & ~ & 1/13/26 & ~\\ 
    %\multicolumn{2}{|c|}{+/-/=} & 0/8/32 &       & 0/15/25 &  \\
  \bottomrule
\end{tabular}
\end{center}
\end{footnotesize}{$+$, $-$, and $=$ denote the performance of the selected algorithm is significantly better, worse, and equivalent to the corresponding variant, respectively.}
\end{table*}

   \begin{table*}%[htbp]
   \begin{footnotesize}
   \begin{center}
  \caption{Comparison results on IGD metric (median and IQR) for C-MOEA/D and C-TAEA against their corresponding variants on C-DTLZ and DC-DTLZ synthetic problems}
  \label{tab:DTLZIGDb}
  \centering
  \begin{tabular}{lllllll}
    \toprule
        Problem & M & $v$C-MOEA/D & C-MOEA/D & $v$C-TAEA & C-TAEA \\
    \midrule
    C1\_DTLZ1 & 2     & 1.9806e-3 (6.63e-5) = & 1.9983e-3 (6.75e-5) & 2.0463e-3 (2.03e-4) = & 2.0635e-3 (1.42e-4) \\
    C1\_DTLZ3 & 2     & 4.3695e-3 (6.70e-5) = & 4.3722e-3 (2.39e-5) & 4.4807e-3 (2.96e-5) = & 4.5077e-3 (1.49e-4) \\
    C2\_DTLZ2 & 2     & 4.5934e-2 (6.52e-6) + & 4.5938e-2 (2.94e-3) & 4.6560e-2 (5.70e-4) = & 4.6475e-2 (5.21e-4) \\
    C3\_DTLZ4 & 2     & 8.6264e-3 (1.37e-4) = & 8.6226e-3 (1.43e-4) & 2.3844e-2 (9.46e-3) = & 2.3972e-2 (7.43e-3) \\
    DC1\_DTLZ1 & 2     & 6.7303e-2 (1.95e-2) = & 4.7832e-2 (1.95e-2) & 4.7710e-2 (1.57e-4) = & 4.7690e-2 (8.25e-5) \\
    DC1\_DTLZ3 & 2     & 1.1861e-1 (1.47e-4) = & 1.1858e-1 (1.05e-4) & 1.2002e-1 (1.11e-3) = & 1.2023e-1 (1.27e-3) \\
    DC2\_DTLZ1 & 2     & 1.9916e-3 (6.11e-4) = & 2.0449e-3 (1.95e-1) & 1.9973e-3 (3.11e-5) = & 1.9926e-3 (9.46e-6) \\
    DC2\_DTLZ3 & 2     & 5.5570e-1 (1.89e-4) - & 5.5566e-1 (4.43e-5) & 4.5447e-3 (3.08e-4) = & 4.5825e-3 (1.72e-4) \\
    DC3\_DTLZ1 & 2     & 5.3285e-2 (1.73e-1) = & 5.3109e-2 (1.73e-1) & 5.3110e-2 (2.35e-4) = & 5.3138e-2 (1.87e-4) \\
    DC3\_DTLZ3 & 2     & 1.2466e+0 (4.91e-1) = & 1.2466e+0 (6.55e-1) & 1.1992e-1 (1.04e-3) = & 1.1992e-1 (1.06e-3) \\
    C1\_DTLZ1 & 3     & 2.0393e-2 (1.37e-4) = & 2.0351e-2 (1.23e-4) & 2.2863e-2 (4.86e-4) = & 2.3028e-2 (4.29e-4) \\
    C1\_DTLZ3 & 3     & 8.0109e+0 (7.89e+0) = & 8.0081e+0 (7.96e+0) & 7.0166e-2 (1.46e-2) = & 7.1967e-2 (3.41e-2) \\
    C2\_DTLZ2 & 3     & 1.1109e-1 (3.70e-4) + & 1.1171e-1 (6.95e-3) & 1.2016e-1 (2.02e-3) = & 1.1968e-1 (1.55e-3) \\
    C3\_DTLZ4 & 3     & 9.1424e-2 (5.26e-5) = & 9.1438e-2 (6.92e-5) & 1.1189e-1 (3.31e-3) = & 1.1176e-1 (2.25e-3) \\
    DC1\_DTLZ1 & 3     & 4.6443e-2 (1.53e-2) = & 4.6445e-2 (8.69e-5) & 4.9526e-2 (6.40e-4) = & 4.9587e-2 (5.07e-4) \\
    DC1\_DTLZ3 & 3     & 1.1878e-1 (3.34e-5) = & 1.1877e-1 (5.22e-5) & 1.3160e-1 (3.14e-3) = & 1.3168e-1 (3.53e-3) \\
    DC2\_DTLZ1 & 3     & 2.1047e-2 (1.44e-1) = & 1.6419e-1 (1.44e-1) & 2.3140e-2 (2.63e-4) = & 2.3105e-2 (2.05e-4) \\
    DC2\_DTLZ3 & 3     & 5.6157e-1 (2.93e-3) = & 5.6085e-1 (6.61e-3) & 6.3613e-2 (6.06e-3) = & 6.7598e-2 (5.04e-1) \\
    DC3\_DTLZ1 & 3     & 9.4843e-2 (1.14e-1) = & 1.8752e-1 (1.15e-1) & 7.5216e-2 (7.42e-4) = & 7.5670e-2 (9.28e-4) \\
    DC3\_DTLZ3 & 3     & 1.2584e+0 (6.50e-1) = & 1.2584e+0 (6.50e-1) & 1.7930e-1 (2.92e-3) = & 1.7887e-1 (1.94e-3) \\
    C1\_DTLZ1 & 5     & 5.2242e-2 (3.73e-4) = & 5.2050e-2 (3.63e-4) & 5.9171e-2 (6.38e-4) + & 5.9498e-2 (6.21e-4) \\
    C1\_DTLZ3 & 5     & 1.1539e+1 (7.63e+0) = & 1.1540e+1 (9.14e-3) & 1.1618e+1 (1.13e+1) = & 1.1624e+1 (9.99e+0) \\
    C2\_DTLZ2 & 5     & 4.1980e-1 (1.62e-1) + & 9.4289e-1 (5.12e-1) & 3.5070e-1 (8.85e-2) = & 3.5156e-1 (8.93e-2) \\
    C3\_DTLZ4 & 5     & 2.4357e-1 (5.53e-5) = & 2.4358e-1 (2.79e-5) & 2.9094e-1 (3.47e-3) = & 2.9051e-1 (3.05e-3) \\
    DC1\_DTLZ1 & 5     & 6.8456e-2 (1.31e-4) = & 6.8449e-2 (3.40e-2) & 6.5322e-2 (4.69e-4) = & 6.5317e-2 (4.95e-4) \\
    DC1\_DTLZ3 & 5     & 1.8897e-1 (7.24e-5) = & 1.8896e-1 (6.87e-5) & 2.7075e-1 (4.25e-2) = & 2.8209e-1 (5.42e-2) \\
    DC2\_DTLZ1 & 5     & 5.3404e-2 (1.05e-1) = & 1.5739e-1 (1.17e-3) & 6.0277e-2 (8.14e-4) = & 6.0367e-2 (7.47e-4) \\
    DC2\_DTLZ3 & 5     & 5.9760e-1 (1.57e-2) = & 5.9703e-1 (5.15e-3) & 1.8681e-1 (3.11e-3) = & 1.8694e-1 (2.80e-3) \\
    DC3\_DTLZ1 & 5     & 1.7090e-1 (1.35e-2) = & 1.7091e-1 (1.67e-2) & 1.5002e-1 (1.08e-2) = & 1.4928e-1 (8.81e-3) \\
    DC3\_DTLZ3 & 5     & 1.3857e+0 (6.58e-1) = & 1.4789e+0 (7.00e-1) & 3.1075e-1 (2.03e-2) = & 3.0741e-1 (3.10e-2) \\
    C1\_DTLZ1 & 10    & 1.0856e-1 (5.08e-4) = & 1.0859e-1 (4.58e-4) & 1.4055e-1 (3.73e-3) = & 1.3974e-1 (1.63e-3) \\
    C1\_DTLZ3 & 10    & 1.4124e+1 (5.61e-3) = & 1.4124e+1 (5.30e-3) & 1.4245e+1 (9.35e-3) = & 1.4245e+1 (1.29e-2) \\
    C2\_DTLZ2 & 10    & 1.0974e+0 (5.17e-1) = & 1.0975e+0 (2.95e-5) & 5.9225e-1 (1.32e-3) = & 5.9207e-1 (1.29e-3) \\
    C3\_DTLZ4 & 10    & 5.6428e-1 (5.94e-4) = & 5.6424e-1 (6.44e-4) & 6.2178e-1 (5.89e-3) = & 6.2454e-1 (5.83e-3) \\
    DC1\_DTLZ1 & 10    & 1.1010e-1 (7.91e-3) = & 1.1006e-1 (1.00e-3) & 2.0132e-1 (5.42e-2) = & 1.8486e-1 (7.41e-2) \\
    DC1\_DTLZ3 & 10    & 4.2192e-1 (4.85e-4) = & 4.2198e-1 (3.32e-4) & 6.1673e-1 (7.38e-2) = & 6.1214e-1 (7.66e-2) \\
    DC2\_DTLZ1 & 10    & 1.7172e-1 (6.50e-2) = & 1.7251e-1 (6.51e-2) & 1.3622e-1 (6.29e-3) + & 1.3868e-1 (5.40e-3) \\
    DC2\_DTLZ3 & 10    & 7.8057e-1 (4.20e-4) = & 7.7857e-1 (4.89e-3) & 4.7892e-1 (3.68e-1) = & 5.1766e-1 (2.93e-1) \\
    DC3\_DTLZ1 & 10    & 2.4250e-1 (3.72e-2) = & 2.4444e-1 (3.83e-2) & 3.9545e+1 (8.63e+1) = & 2.2400e+1 (5.47e+1) \\
    DC3\_DTLZ3 & 10    & 1.9804e+0 (6.96e-1) = & 1.9478e+0 (6.80e-1) & 6.6145e-1 (7.20e-2) = & 6.3789e-1 (5.22e-2) \\
        +/-/= &   & 3/1/36 & ~ & 2/0/38 & ~\\ 
  \bottomrule
\end{tabular}
\end{center}
\end{footnotesize}{$+$, $-$, and $=$ denote the performance of the selected algorithm is significantly better, worse, and equivalent to the corresponding variant, respectively.}
%\end{footnotesize}
\end{table*}

   \begin{table*}%[htbp]
   \begin{footnotesize}
   \begin{center}
  \caption{Comparison results on IGD$^+$ metric (median and IQR) for C-NSGA-II and C-NSGA-III against their corresponding variants on C-DTLZ and DC-DTLZ synthetic problems}
  \label{tab:DTLZIGDplusa}
  \centering
  \begin{tabular}{lllllll}
    \toprule
    Problem & M     & $v$C-NSGA-II & C-NSGA-II & $v$C-NSGA-III & C-NSGA-III \\
    \midrule
    C1\_DTLZ1 & 2     & 5.0000e+2 (0.00e+0) - & 1.8926e-3 (3.10e-4) & 5.0000e+2 (0.00e+0) - & 1.5804e-3 (2.32e-4) \\
    C1\_DTLZ3 & 2     & 2.5233e-3 (5.00e+0) + & 5.0001e+0 (5.00e+0) & 2.2115e-3 (1.21e-4) = & 2.2495e-3 (1.93e-4) \\
    C2\_DTLZ2 & 2     & 6.8663e-2 (1.83e-1) = & 1.8989e-2 (1.10e-1) & 2.3820e-1 (1.10e-1) - & 1.9593e-2 (8.11e-5) \\
    C3\_DTLZ4 & 2     & 6.4722e-3 (4.71e-4) = & 6.4873e-3 (3.05e-4) & 5.4858e-3 (1.50e-4) = & 5.4980e-3 (2.91e-4) \\
    DC1\_DTLZ1 & 2     & 3.3902e-2 (2.76e-2) = & 4.7734e-2 (2.76e-2) & 3.4225e-2 (1.11e-2) = & 3.4052e-2 (1.41e-2) \\
    DC1\_DTLZ3 & 2     & 7.8346e-2 (4.45e-4) = & 7.8243e-2 (5.46e-4) & 7.8949e-2 (4.63e-4) = & 7.8848e-2 (2.98e-4) \\
    DC2\_DTLZ1 & 2     & 5.0000e+2 (0.00e+0) = & 5.0000e+2 (0.00e+0)  & 5.0000e+2 (0.00e+0) - & 1.9646e-1 (2.21e-4) \\
    DC2\_DTLZ3 & 2     & 5.0000e+2 (0.00e+0) = & 5.0000e+2 (0.00e+0)  & 5.0000e+2 (0.00e+0) = & 5.0000e+2 (0.00e+0) \\
    DC3\_DTLZ1 & 2     & 2.2557e-1 (1.82e-1) = & 2.2557e-1 (1.88e-1) & 2.2559e-1 (1.88e-1) = & 2.2559e-1 (1.88e-1) \\
    DC3\_DTLZ3 & 2     & 1.9096e+0 (6.63e-1) = & 1.9096e+0 (6.63e-1) & 1.2470e+0 (1.16e+0) = & 1.2470e+0 (2.04e-4) \\
    C1\_DTLZ1 & 3     & 5.0000e+2 (0.00e+0) - & 1.9780e-2 (1.23e-3) & 5.0000e+2 (0.00e+0) - & 1.4816e-2 (4.68e-4) \\
    C1\_DTLZ3 & 3     & 8.0231e+0 (4.86e-3) = & 8.0244e+0 (5.58e+0) & 1.1572e-1 (7.99e+0) = & 8.0070e+0 (7.99e+0) \\
    C2\_DTLZ2 & 3     & 7.1207e-1 (5.00e+2) - & 6.6972e-2 (1.50e-3) & 2.4909e-1 (1.18e-1) - & 6.4235e-2 (4.59e-3) \\
    C3\_DTLZ4 & 3     & 8.9933e-2 (7.89e-3) = & 9.1550e-2 (8.31e-3) & 4.8326e-2 (1.13e-3) = & 4.8552e-2 (1.73e-3) \\
    DC1\_DTLZ1 & 3     & 3.5345e-2 (1.19e-2) = & 4.5590e-2 (1.24e-2) & 3.4814e-2 (1.34e-2) = & 3.4650e-2 (1.31e-2) \\
    DC1\_DTLZ3 & 3     & 6.2222e-2 (1.27e-3) = & 6.2063e-2 (2.23e-3) & 5.8869e-2 (6.11e-4) = & 5.9110e-2 (6.82e-4) \\
    DC2\_DTLZ1 & 3     & 5.0000e+2 (0.00e+0) = & 5.0000e+2 (0.00e+0) & 5.0000e+2 (0.00e+0) - & 1.6428e-1 (5.39e-4) \\
    DC2\_DTLZ3 & 3     & 5.0000e+2 (0.00e+0) = & 5.0000e+2 (0.00e+0) & 5.0000e+2 (0.00e+0) = & 5.0000e+2 (0.00e+0) \\
    DC3\_DTLZ1 & 3     & 1.8801e-1 (7.99e-4) = & 1.8791e-1 (8.65e-2) & 1.8981e-1 (1.32e-1) = & 1.9050e-1 (1.59e-3) \\
    DC3\_DTLZ3 & 3     & 1.2630e+0 (6.62e-1) = & 1.2641e+0 (6.61e-1) & 1.2642e+0 (2.45e-3) = & 1.2634e+0 (3.07e-3) \\
    C1\_DTLZ1 & 5     & 5.0000e+2 (0.00e+0) - & 4.1865e-2 (8.30e-4) & 5.0000e+2 (0.00e+0) - & 3.7164e-2 (2.23e-4) \\
    C1\_DTLZ3 & 5     & 1.1636e+1 (2.70e-2) = & 1.1641e+1 (2.35e-2) & 1.1551e+1 (1.03e+1) = & 1.1553e+1 (1.14e+1) \\
    C2\_DTLZ2 & 5     & 5.0000e+2 (0.00e+0) - & 1.8504e-1 (4.57e-1) & 5.0000e+2 (0.00e+0) - & 6.4091e-1 (4.60e-1) \\
    C3\_DTLZ4 & 5     & 3.6673e-1 (1.71e-2) = & 3.6634e-1 (3.36e-2) & 1.2526e-1 (5.86e-3) = & 1.2538e-1 (7.06e-3) \\
    DC1\_DTLZ1 & 5     & 5.7125e-2 (2.81e-3) = & 5.7273e-2 (2.99e-3) & 4.4933e-2 (2.58e-3) - & 4.4788e-2 (3.66e-4) \\
    DC1\_DTLZ3 & 5     & 1.2280e-1 (1.02e-2) = & 1.2461e-1 (9.92e-3) & 8.0870e-2 (5.46e-4) = & 8.0828e-2 (5.21e-4) \\
    DC2\_DTLZ1 & 5     & 5.0000e+2 (0.00e+0) = & 5.0000e+2 (0.00e+0) & 5.0000e+2 (0.00e+0) - & 1.5770e-1 (1.52e-3) \\
    DC2\_DTLZ3 & 5     & 5.0000e+2 (0.00e+0) = & 5.0000e+2 (0.00e+0) & 5.0000e+2 (0.00e+0) = & 5.0000e+2 (0.00e+0) \\
    DC3\_DTLZ1 & 5     & 7.7009e-2 (9.55e-2) = & 1.7136e-1 (9.63e-2) & 1.6812e-1 (9.37e-2) = & 1.6977e-1 (9.57e-2) \\
    DC3\_DTLZ3 & 5     & 6.6963e-1 (4.85e-1) = & 6.6757e-1 (1.23e-2) & 6.5197e-1 (6.54e-1) - & 6.5033e-1 (6.72e-3) \\
    C1\_DTLZ1 & 10    & 5.0000e+2 (0.00e+0) - & 7.8923e-2 (2.64e-3) & 5.0000e+2 (0.00e+0) - & 6.8997e-2 (4.37e-4) \\
    C1\_DTLZ3 & 10    & 1.0343e+2 (3.83e+1) = & 8.7374e+1 (2.41e+1) & 1.7425e+0 (1.40e+1) = & 1.2128e+0 (1.40e+1) \\
    C2\_DTLZ2 & 10    & 5.0000e+2 (0.00e+0) - & 2.8210e-1 (4.99e-1) & 5.0000e+2 (0.00e+0) - & 7.7672e-1 (5.02e-1) \\
    C3\_DTLZ4 & 10    & 2.0327e+0 (1.36e-1) = & 2.0190e+0 (8.04e-2) & 2.8325e-1 (3.60e-3) = & 2.8253e-1 (2.50e-3) \\
    DC1\_DTLZ1 & 10    & 5.7229e+0 (3.68e+0) = & 4.6930e+0 (2.95e+0) & 7.4834e-2 (1.31e-2) = & 7.3667e-2 (1.04e-2) \\
    DC1\_DTLZ3 & 10    & 5.2779e+1 (5.52e+0) = & 5.3486e+1 (6.11e+0) & 3.0012e-1 (9.94e-2) = & 3.0257e-1 (1.19e-1) \\
    DC2\_DTLZ1 & 10    & 5.0000e+2 (0.00e+0) = & 5.0000e+2 (0.00e+0) & 5.0000e+2 (0.00e+0) - & 1.6650e-1 (7.41e-3) \\
    DC2\_DTLZ3 & 10    & 5.0000e+2 (0.00e+0) = & 5.0000e+2 (0.00e+0) & 5.0000e+2 (0.00e+0) - & 5.0000e+2 (4.99e+2) \\
    DC3\_DTLZ1 & 10    & 1.2949e-1 (1.50e-1) = & 1.3183e-1 (9.79e-2) & 1.2768e-1 (9.72e-2) = & 1.2771e-1 (1.44e-2) \\
    DC3\_DTLZ3 & 10    & 4.7120e+1 (1.85e+1) = & 4.6954e+1 (2.01e+1) & 1.4004e+0 (7.04e-1) = & 1.4036e+0 (1.16e-1) \\
    +/-/= &   & 1/7/32 & ~ & 1/15/25 & ~\\ 
    %\multicolumn{2}{|c|}{+/-/=} & 0/8/32 &       & 0/15/25 &  \\
  \bottomrule
\end{tabular}
\end{center}
\end{footnotesize}{$+$, $-$, and $=$ denote the performance of the selected algorithm is significantly better, worse, and equivalent to the corresponding variant, respectively.}
\end{table*}

   \begin{table*}%[htbp]
   \begin{footnotesize}
   \begin{center}
  \caption{Comparison results on IGD$^+$ metric (median and IQR) for C-MOEA/D and C-TAEA against their corresponding variants on C-DTLZ and DC-DTLZ synthetic problems}
  \label{tab:DTLZIGDplusb}
  \centering
  \begin{tabular}{lllllll}
    \toprule
        Problem & M & $v$C-MOEA/D & C-MOEA/D & $v$C-TAEA & C-TAEA \\
    \midrule
    C1\_DTLZ1 & 2     & 1.5310e-3 (2.17e-4) = & 1.5556e-3 (1.43e-4) & 1.5918e-3 (2.58e-4) = & 1.5459e-3 (1.28e-4) \\
    C1\_DTLZ3 & 2     & 2.1864e-3 (1.98e-4) = & 2.1826e-3 (1.88e-4) & 2.2768e-3 (1.74e-4) = & 2.3241e-3 (1.72e-4) \\
    C2\_DTLZ2 & 2     & 1.9318e-2 (5.02e-6) = & 1.9320e-2 (1.14e-3) & 1.9375e-2 (1.40e-4) = & 1.9365e-2 (1.86e-4) \\
    C3\_DTLZ4 & 2     & 4.9705e-3 (9.62e-5) = & 4.9878e-3 (8.83e-5) & 1.6502e-2 (4.68e-3) = & 1.6464e-2 (5.08e-3) \\
    DC1\_DTLZ1 & 2     & 4.7606e-2 (1.38e-2) = & 3.3926e-2 (1.38e-2) & 3.3778e-2 (1.45e-4) = & 3.3755e-2 (9.64e-5) \\
    DC1\_DTLZ3 & 2     & 7.8577e-2 (2.73e-4) = & 7.8549e-2 (1.71e-4) & 7.8927e-2 (5.32e-4) = & 7.9107e-2 (5.67e-4) \\
    DC2\_DTLZ1 & 2     & 1.5123e-3 (1.04e-3) = & 1.6671e-3 (1.95e-1) & 1.4562e-3 (1.64e-4) = & 1.4311e-3 (5.98e-5) \\
    DC2\_DTLZ3 & 2     & 5.8726e-1 (4.99e+2) = & 5.0000e+2 (4.99e+2) & 2.4093e-3 (3.72e-4) = & 2.4582e-3 (5.17e-4) \\
    DC3\_DTLZ1 & 2     & 3.8125e-2 (1.88e-1) = & 3.7760e-2 (1.88e-1) & 3.7585e-2 (1.52e-4) = & 3.7597e-2 (1.46e-4) \\
    DC3\_DTLZ3 & 2     & 1.2465e+0 (4.93e-1) = & 1.2465e+0 (6.58e-1) & 7.8679e-2 (3.92e-4) = & 7.8725e-2 (4.00e-4) \\
    C1\_DTLZ1 & 3     & 1.4584e-2 (1.48e-4) = & 1.4619e-2 (1.51e-4) & 1.6366e-2 (4.58e-4) = & 1.6313e-2 (4.79e-4) \\
    C1\_DTLZ3 & 3     & 8.0108e+0 (7.91e+0) = & 8.0080e+0 (7.99e+0) & 3.0140e-2 (6.03e-3) = & 3.0807e-2 (8.86e-3) \\
    C2\_DTLZ2 & 3     & 6.0643e-2 (7.56e-5) + & 6.0812e-2 (3.52e-3) & 6.3992e-2 (6.25e-4) = & 6.3799e-2 (6.01e-4) \\
    C3\_DTLZ4 & 3     & 4.5739e-2 (1.95e-4) + & 4.5877e-2 (2.05e-4) & 6.1159e-2 (2.35e-3) = & 6.1124e-2 (2.22e-3) \\
    DC1\_DTLZ1 & 3     & 3.0815e-2 (9.68e-3) = & 3.0816e-2 (1.80e-4) & 3.2767e-2 (4.51e-4) = & 3.2778e-2 (3.93e-4) \\
    DC1\_DTLZ3 & 3     & 5.7116e-2 (1.31e-4) = & 5.7107e-2 (1.23e-4) & 6.1142e-2 (8.37e-4) = & 6.0918e-2 (1.73e-3) \\
    DC2\_DTLZ1 & 3     & 1.6339e-2 (1.49e-1) = & 1.6419e-1 (1.50e-1) & 1.6325e-2 (2.56e-4) = & 1.6300e-2 (2.73e-4) \\
    DC2\_DTLZ3 & 3     & 5.6442e-1 (4.99e+2) = & 5.0000e+2 (4.99e+2) & 2.9538e-2 (1.05e-2) = & 3.3519e-2 (5.39e-1) \\
    DC3\_DTLZ1 & 3     & 6.8650e-2 (1.36e-1) = & 1.8691e-1 (1.35e-1) & 5.2220e-2 (7.47e-4) = & 5.2672e-2 (8.20e-4) \\
    DC3\_DTLZ3 & 3     & 1.2581e+0 (6.55e-1) = & 1.2581e+0 (6.55e-1) & 1.0373e-1 (9.03e-4) = & 1.0361e-1 (5.89e-4) \\
    C1\_DTLZ1 & 5     & 3.7159e-2 (2.42e-4) = & 3.7082e-2 (2.16e-4) & 4.1463e-2 (4.51e-4) = & 4.1676e-2 (3.73e-4) \\
    C1\_DTLZ3 & 5     & 1.1539e+1 (7.64e+0) = & 1.1539e+1 (9.06e-3) & 1.1614e+1 (1.15e+1) = & 1.1621e+1 (1.00e+1) \\
    C2\_DTLZ2 & 5     & 1.8046e-1 (7.94e-2) + & 6.4092e-1 (4.12e-1) & 1.7088e-1 (1.29e-2) = & 1.7154e-1 (1.30e-2) \\
    C3\_DTLZ4 & 5     & 1.0998e-1 (2.03e-4) = & 1.1000e-1 (2.02e-4) & 1.4653e-1 (2.53e-3) = & 1.4578e-1 (2.85e-3) \\
    DC1\_DTLZ1 & 5     & 4.4963e-2 (3.71e-4) = & 4.4931e-2 (3.23e-2) & 4.4716e-2 (4.58e-4) = & 4.4667e-2 (4.05e-4) \\
    DC1\_DTLZ3 & 5     & 7.4896e-2 (8.78e-5) = & 7.4903e-2 (9.00e-5) & 1.0641e-1 (1.21e-2) = & 1.1172e-1 (2.01e-2) \\
    DC2\_DTLZ1 & 5     & 4.0201e-2 (1.19e-1) = & 1.5712e-1 (1.11e-3) & 4.2945e-2 (2.07e-3) = & 4.2822e-2 (3.71e-4) \\
    DC2\_DTLZ3 & 5     & 5.0000e+2 (4.99e+2) - & 5.0000e+2 (0.00e+0) & 7.2326e-2 (2.11e-3) = & 7.2640e-2 (5.41e-3) \\
    DC3\_DTLZ1 & 5     & 1.6621e-1 (8.24e-3) = & 1.6621e-1 (5.87e-2) & 1.0248e-1 (8.26e-3) = & 9.9647e-2 (6.08e-3) \\
    DC3\_DTLZ3 & 5     & 1.3759e+0 (6.59e-1) = & 1.4747e+0 (7.02e-1) & 1.5547e-1 (6.27e-3) = & 1.5492e-1 (1.14e-2) \\
    C1\_DTLZ1 & 10    & 6.9204e-2 (2.81e-4) = & 6.9257e-2 (2.82e-4) & 1.0257e-1 (4.01e-3) = & 1.0186e-1 (1.67e-3) \\
    C1\_DTLZ3 & 10    & 1.4122e+1 (5.12e-3) = & 1.4122e+1 (5.23e-3) & 1.4234e+1 (8.83e-3) = & 1.4234e+1 (1.13e-2) \\
    C2\_DTLZ2 & 10    & 7.7671e-1 (5.15e-1) = & 7.7671e-1 (1.00e-7) & 2.7327e-1 (6.94e-4) = & 2.7317e-1 (8.11e-4) \\
    C3\_DTLZ4 & 10    & 2.7331e-1 (1.15e-4) = & 2.7330e-1 (1.34e-4) & 2.9759e-1 (3.70e-3) = & 2.9821e-1 (2.71e-3) \\
    DC1\_DTLZ1 & 10    & 6.9459e-2 (1.50e-3) = & 6.9435e-2 (1.38e-3) & 1.3753e-1 (4.41e-2) = & 1.2291e-1 (5.45e-2) \\
    DC1\_DTLZ3 & 10    & 1.7376e-1 (2.41e-4) = & 1.7378e-1 (1.74e-4) & 3.7934e-1 (1.27e-1) = & 3.8127e-1 (1.20e-1) \\
    DC2\_DTLZ1 & 10    & 1.6142e-1 (9.35e-2) = & 1.6193e-1 (9.32e-2) & 9.9801e-2 (6.27e-3) + & 1.0146e-1 (5.48e-3) \\
    DC2\_DTLZ3 & 10    & 5.0000e+2 (0.00e+0) = & 5.0000e+2 (0.00e+0) & 2.1134e-1 (4.89e-1) = & 2.2904e-1 (2.81e-1) \\
    DC3\_DTLZ1 & 10    & 2.0318e-1 (2.18e-2) = & 2.0511e-1 (2.19e-2) & 9.4528e+1 (4.84e+2) = & 1.4031e+2 (4.83e+2) \\
    DC3\_DTLZ3 & 10    & 1.8928e+0 (7.14e-1) = & 1.8589e+0 (7.09e-1) & 3.4529e-1 (5.01e-2) = & 3.2454e-1 (3.55e-2) \\
        +/-/= &   & 3/1/36 & ~ & 1/0/39 & ~\\ 
  \bottomrule
\end{tabular}
\end{center}
\end{footnotesize}{$+$, $-$, and $=$ denote the performance of the selected algorithm is significantly better, worse, and equivalent to the corresponding variant, respectively.}
%\end{footnotesize}
\end{table*}

   \begin{table*}%[htbp]
   \begin{footnotesize}
   \begin{center}
  \caption{Comparison results on HV metric (median and IQR) for C-NSGA-II and C-NSGA-III against their corresponding variants on C-DTLZ and DC-DTLZ synthetic problems}
  \label{tab:DTLZHVa}
  \centering
  \begin{tabular}{lllllll}
    \toprule
    Problem & M     & $v$C-NSGA-II & C-NSGA-II & $v$C-NSGA-III & C-NSGA-III \\
    \midrule
    C1\_DTLZ1 & 2     & 0.0000e+0 (0.00e+0) - & 5.8011e-1 (1.55e-3) & 0.0000e+0 (0.00e+0) - & 5.8114e-1 (1.35e-3) \\
    C1\_DTLZ3 & 2     & 3.4627e-1 (3.47e-1) = & 0.0000e+0 (3.46e-1) & 3.4662e-1 (1.96e-4) = & 3.4656e-1 (3.05e-4) \\
    C2\_DTLZ2 & 2     & 2.6263e-1 (1.14e-1) - & 3.1476e-1 (5.22e-2) & 2.1045e-1 (5.21e-2) - & 3.1351e-1 (1.94e-4) \\
    C3\_DTLZ4 & 2     & 5.3696e-1 (3.77e-4) = & 5.3695e-1 (2.86e-4) & 5.3775e-1 (1.34e-4) = & 5.3773e-1 (2.14e-4) \\
    DC1\_DTLZ1 & 2     & 4.7470e-1 (9.12e-2) = & 4.4428e-1 (9.12e-2) & 4.7364e-1 (2.52e-2) = & 4.7420e-1 (3.14e-2) \\
    DC1\_DTLZ3 & 2     & 2.5499e-1 (7.11e-4) = & 2.5515e-1 (8.41e-4) & 2.5375e-1 (7.40e-4) = & 2.5389e-1 (5.79e-4) \\
    DC2\_DTLZ1 & 2     & 0.0000e+0 (0.00e+0) = & 0.0000e+0 (0.00e+0) & 0.0000e+0 (0.00e+0) - & 1.6704e-1 (3.07e-4) \\
    DC2\_DTLZ3 & 2     & 0.0000e+0 (0.00e+0) = & 0.0000e+0 (0.00e+0) & 0.0000e+0 (0.00e+0) - & 0.0000e+0 (0.00e+0) \\
    DC3\_DTLZ1 & 2     & 1.0190e-1 (3.45e-1) = & 1.0186e-1 (3.63e-1) & 1.0166e-1 (3.62e-1) = & 1.0163e-1 (3.62e-1) \\
    DC3\_DTLZ3 & 2     & 0.0000e+0 (0.00e+0) = & 0.0000e+0 (0.00e+0) & 0.0000e+0 (0.00e+0) = & 0.0000e+0 (0.00e+0) \\
    C1\_DTLZ1 & 3     & 0.0000e+0 (0.00e+0) - & 8.1430e-1 (5.06e-3) & 0.0000e+0 (0.00e+0) - & 8.3432e-1 (1.04e-2) \\
    C1\_DTLZ3 & 3     & 0.0000e+0 (0.00e+0) = & 0.0000e+0 (4.55e-3) & 4.0675e-1 (5.58e-1) = & 0.0000e+0 (5.56e-1) \\
    C2\_DTLZ2 & 3     & 0.0000e+0 (2.84e-1) - & 4.5244e-1 (3.00e-3) & 2.8529e-1 (1.07e-1) - & 4.5633e-1 (1.22e-2) \\
    C3\_DTLZ4 & 3     & 7.6315e-1 (5.57e-3) = & 7.6145e-1 (6.04e-3) & 7.9393e-1 (6.80e-4) = & 7.9385e-1 (1.13e-3) \\
    DC1\_DTLZ1 & 3     & 7.7700e-1 (3.76e-2) = & 7.5391e-1 (3.53e-2) & 7.9400e-1 (4.19e-2) = & 7.9462e-1 (4.15e-2) \\
    DC1\_DTLZ3 & 3     & 4.8795e-1 (3.61e-3) = & 4.8825e-1 (3.82e-3) & 4.9565e-1 (1.21e-3) = & 4.9525e-1 (1.08e-3) \\
    DC2\_DTLZ1 & 3     & 0.0000e+0 (0.00e+0) = & 0.0000e+0 (0.00e+0) & 0.0000e+0 (0.00e+0) - & 4.7822e-1 (1.29e-3) \\
    DC2\_DTLZ3 & 3     & 0.0000e+0 (0.00e+0) = & 0.0000e+0 (0.00e+0) & 0.0000e+0 (0.00e+0) = & 0.0000e+0 (0.00e+0) \\
    DC3\_DTLZ1 & 3     & 3.3159e-1 (3.09e-3) = & 3.3027e-1 (1.89e-1) & 3.2286e-1 (3.02e-1) = & 3.2313e-1 (6.61e-3) \\
    DC3\_DTLZ3 & 3     & 0.0000e+0 (0.00e+0) = & 0.0000e+0 (0.00e+0) & 0.0000e+0 (0.00e+0) = & 0.0000e+0 (0.00e+0) \\
    C1\_DTLZ1 & 5     & 0.0000e+0 (0.00e+0) - & 9.6888e-1 (7.20e-3) & 0.0000e+0 (0.00e+0) - & 9.7462e-1 (5.94e-3) \\
    C1\_DTLZ3 & 5     & 0.0000e+0 (0.00e+0) = & 0.0000e+0 (0.00e+0) & 0.0000e+0 (0.00e+0) = & 0.0000e+0 (6.25e-1) \\
    C2\_DTLZ2 & 5     & 0.0000e+0 (0.00e+0) - & 4.3336e-1 (3.26e-1) & 0.0000e+0 (0.00e+0) - & 1.0907e-1 (3.29e-1) \\
    C3\_DTLZ4 & 5     & 8.6625e-1 (1.43e-2) = & 8.6589e-1 (1.63e-2) & 9.6342e-1 (1.08e-3) = & 9.6343e-1 (1.45e-3) \\
    DC1\_DTLZ1 & 5     & 9.5581e-1 (7.03e-3) = & 9.5535e-1 (3.15e-3) & 9.7500e-1 (4.09e-3) - & 9.7530e-1 (2.79e-4) \\
    DC1\_DTLZ3 & 5     & 6.8920e-1 (2.39e-2) = & 6.8676e-1 (1.35e-2) & 7.7856e-1 (1.46e-3) = & 7.7832e-1 (1.03e-3) \\
    DC2\_DTLZ1 & 5     & 0.0000e+0 (0.00e+0) = & 0.0000e+0 (0.00e+0) & 0.0000e+0 (0.00e+0) - & 8.3132e-1 (3.31e-3) \\
    DC2\_DTLZ3 & 5     & 0.0000e+0 (0.00e+0) = & 0.0000e+0 (0.00e+0) & 0.0000e+0 (0.00e+0) = & 0.0000e+0 (0.00e+0) \\
    DC3\_DTLZ1 & 5     & 6.9404e-1 (2.31e-1) = & 4.6677e-1 (2.31e-1) & 4.7244e-1 (2.25e-1) = & 4.7091e-1 (2.30e-1) \\
    DC3\_DTLZ3 & 5     & 0.0000e+0 (0.00e+0) = & 0.0000e+0 (0.00e+0) & 0.0000e+0 (0.00e+0) = & 0.0000e+0 (0.00e+0) \\
    C1\_DTLZ1 & 10    & 0.0000e+0 (0.00e+0) - & 9.9885e-1 (2.29e-3) & 0.0000e+0 (0.00e+0) - & 9.9548e-1 (7.96e-3) \\
    C1\_DTLZ3 & 10    & 0.0000e+0 (0.00e+0) = & 0.0000e+0 (0.00e+0) & 0.0000e+0 (9.69e-1) = & 1.8135e-3 (9.68e-1) \\
    C2\_DTLZ2 & 10    & 0.0000e+0 (0.00e+0) - & 6.6460e-1 (5.59e-1) & 0.0000e+0 (0.00e+0) - & 1.0909e-1 (5.58e-1) \\
    C3\_DTLZ4 & 10    & 5.6443e-2 (9.06e-3) = & 5.8896e-2 (1.03e-2) & 9.9936e-1 (6.20e-5) = & 9.9937e-1 (4.00e-5) \\
    DC1\_DTLZ1 & 10    & 0.0000e+0 (0.00e+0) = & 0.0000e+0 (0.00e+0) & 9.9936e-1 (1.14e-3) = & 9.9959e-1 (8.12e-4) \\
    DC1\_DTLZ3 & 10    & 0.0000e+0 (0.00e+0) = & 0.0000e+0 (0.00e+0) & 8.1537e-1 (1.35e-1) = & 8.3224e-1 (1.55e-1) \\
    DC2\_DTLZ1 & 10    & 0.0000e+0 (0.00e+0) = & 0.0000e+0 (0.00e+0) & 0.0000e+0 (0.00e+0) - & 9.7609e-1 (3.16e-3) \\
    DC2\_DTLZ3 & 10    & 0.0000e+0 (0.00e+0) = & 0.0000e+0 (0.00e+0) & 0.0000e+0 (0.00e+0) - & 0.0000e+0 (9.92e-2) \\
    DC3\_DTLZ1 & 10    & 6.6025e-1 (4.26e-1) = & 6.4958e-1 (3.13e-1) & 6.8799e-1 (2.25e-1) = & 6.8643e-1 (6.45e-2) \\
    DC3\_DTLZ3 & 10    & 0.0000e+0 (0.00e+0) = & 0.0000e+0 (0.00e+0) & 0.0000e+0 (0.00e+0) = & 0.0000e+0 (0.00e+0) \\
    +/-/= &   & 0/8/32 & ~ & 0/15/25 & ~\\ 
    %\multicolumn{2}{|c|}{+/-/=} & 0/8/32 &       & 0/15/25 &  \\
  \bottomrule
\end{tabular}
\end{center}
\end{footnotesize}{$+$, $-$, and $=$ denote the performance of the selected algorithm is significantly better, worse, and equivalent to the corresponding variant, respectively.}
\end{table*}

   \begin{table*}%[htbp]
   \begin{footnotesize}
   \begin{center}
  \caption{Comparison results on HV metric (median and IQR) for C-MOEA/D and C-TAEA against their corresponding variants on C-DTLZ and DC-DTLZ synthetic problems}
  \label{tab:DTLZHVb}
  \centering
  \begin{tabular}{lllllll}
    \toprule
        Problem & M & $v$C-MOEA/D & C-MOEA/D & $v$C-TAEA & C-TAEA \\
    \midrule        
        C1\_DTLZ1 & 2 & 5.8135e-1 (1.12e-3) = & 5.8111e-1 (1.10e-3) & 5.8115e-1 (1.31e-3) = & 5.8145e-1 (6.87e-4) \\
        C1\_DTLZ3 & 2 & 3.4666e-1 (3.17e-4) = & 3.4667e-1 (3.06e-4) & 3.4647e-1 (2.76e-4) = & 3.4639e-1 (2.71e-4) \\ 
        C2\_DTLZ2 & 2 & 3.1401e-1 (7.03e-6) = & 3.1400e-1 (2.33e-3) & 3.1405e-1 (2.61e-4) = & 3.1406e-1 (3.12e-4) \\
        C3\_DTLZ4 & 2 & 5.3812e-1 (8.77e-5) = & 5.3812e-1 (7.84e-5) & 5.2884e-1 (3.27e-3) = & 5.2876e-1 (3.96e-3) \\
        DC1\_DTLZ1 & 2 & 4.4465e-1 (3.03e-2) = & 4.7461e-1 (3.03e-2) & 4.7510e-1 (4.79e-4) = & 4.7518e-1 (3.18e-4) \\ 
        DC1\_DTLZ3 & 2 & 2.5443e-1 (4.53e-4) = & 2.5453e-1 (2.80e-4) & 2.5372e-1 (8.27e-4) = & 2.5365e-1 (1.04e-3) \\
        DC2\_DTLZ1 & 2 & 5.8178e-1 (3.28e-3) = & 5.8128e-1 (4.15e-1) & 5.8196e-1 (5.37e-4) = & 5.8205e-1 (1.97e-4) \\ 
        DC2\_DTLZ3 & 2 & 0.0000e+0 (0.00e+0) = & 0.0000e+0 (0.00e+0) & 3.4625e-1 (6.81e-4) = & 3.4617e-1 (7.94e-4) \\ 
        DC3\_DTLZ1 & 2 & 4.6290e-1 (3.64e-1) = & 4.6402e-1 (3.64e-1) & 4.6460e-1 (4.98e-4) = & 4.6455e-1 (4.86e-4) \\ 
        DC3\_DTLZ3 & 2 & 0.0000e+0 (0.00e+0) = & 0.0000e+0 (0.00e+0) & 2.5409e-1 (8.31e-4) = & 2.5398e-1 (8.05e-4) \\ 
        C1\_DTLZ1 & 3 & 8.3870e-1 (3.88e-3) = & 8.3791e-1 (4.11e-3) & 8.3378e-1 (6.44e-3) = & 8.3443e-1 (7.65e-3) \\
        C1\_DTLZ3 & 3 & 0.0000e+0 (4.30e-1) = & 0.0000e+0 (5.57e-1) & 5.4214e-1 (1.35e-2) = & 5.4100e-1 (2.05e-2) \\
        C2\_DTLZ2 & 3 & 4.6442e-1 (3.28e-4) + & 4.6393e-1 (6.88e-3) & 4.5725e-1 (1.94e-3) = & 4.5747e-1 (1.22e-3) \\ 
        C3\_DTLZ4 & 3 & 7.9563e-1 (1.16e-4) + & 7.9552e-1 (1.40e-4) & 7.8492e-1 (1.29e-3) = & 7.8495e-1 (1.45e-3) \\
        DC1\_DTLZ1 & 3 & 8.0084e-1 (2.70e-2) = & 8.0083e-1 (3.86e-4) & 8.0162e-1 (9.78e-4) = & 8.0147e-1 (1.18e-3) \\
        DC1\_DTLZ3 & 3 & 4.9964e-1 (2.33e-4) = & 4.9965e-1 (2.27e-4) & 4.8952e-1 (2.85e-3) = & 4.8968e-1 (3.20e-3) \\ 
        DC2\_DTLZ1 & 3 & 8.3782e-1 (3.63e-1) = & 4.7832e-1 (3.64e-1) & 8.3817e-1 (4.76e-4) = & 8.3816e-1 (5.32e-4) \\
        DC2\_DTLZ3 & 3 & 7.5336e-3 (7.98e-3) = & 0.0000e+0 (8.02e-3) & 5.4425e-1 (1.81e-2) = & 5.3677e-1 (5.41e-1) \\
        DC3\_DTLZ1 & 3 & 5.6568e-1 (3.13e-1) = & 3.3316e-1 (3.08e-1) & 6.4184e-1 (2.00e-3) = & 6.4042e-1 (2.36e-3) \\
        DC3\_DTLZ3 & 3 & 0.0000e+0 (0.00e+0) = & 0.0000e+0 (0.00e+0) & 4.1134e-1 (1.88e-3) = & 4.1123e-1 (1.46e-3) \\ 
        C1\_DTLZ1 & 5 & 9.7671e-1 (4.64e-3) + & 9.7229e-1 (5.76e-3) & 9.7733e-1 (2.98e-3) = & 9.7763e-1 (1.39e-3) \\ 
        C1\_DTLZ3 & 5 & 0.0000e+0 (0.00e+0) = & 0.0000e+0 (0.00e+0) & 0.0000e+0 (6.91e-1) = & 0.0000e+0 (0.00e+0) \\ 
        C2\_DTLZ2 & 5 & 4.3847e-1 (1.25e-1) + & 1.0908e-1 (2.86e-1) & 4.9736e-1 (6.59e-2) = & 4.9539e-1 (6.67e-2) \\ 
        C3\_DTLZ4 & 5 & 9.6616e-1 (3.08e-4) = & 9.6620e-1 (2.25e-4) & 9.5702e-1 (6.63e-4) = & 9.5705e-1 (8.68e-4) \\
        DC1\_DTLZ1 & 5 & 9.7203e-1 (5.19e-4) = & 9.7206e-1 (2.01e-1) & 9.7564e-1 (4.13e-4) = & 9.7565e-1 (3.42e-4) \\
        DC1\_DTLZ3 & 5 & 7.8985e-1 (5.97e-4) = & 7.8995e-1 (6.85e-4) & 7.2271e-1 (3.21e-2) = & 7.1022e-1 (5.24e-2) \\
        DC2\_DTLZ1 & 5 & 9.7722e-1 (1.46e-1) = & 8.3356e-1 (1.83e-3) & 9.7725e-1 (7.72e-4) = & 9.7728e-1 (3.16e-4) \\
        DC2\_DTLZ3 & 5 & 0.0000e+0 (6.82e-2) - & 0.0000e+0 (0.00e+0) & 7.9193e-1 (4.27e-3) = & 7.9119e-1 (9.77e-3) \\
        DC3\_DTLZ1 & 5 & 4.2684e-1 (2.85e-2) = & 4.2638e-1 (1.54e-1) & 6.4737e-1 (1.37e-2) = & 6.5161e-1 (1.74e-2) \\
        DC3\_DTLZ3 & 5 & 0.0000e+0 (0.00e+0) = & 0.0000e+0 (0.00e+0) & 6.1990e-1 (1.67e-2) = & 6.1723e-1 (2.91e-2) \\
        C1\_DTLZ1 & 10 & 9.9135e-1 (7.88e-3) = & 9.9222e-1 (9.43e-3) & 9.9929e-1 (7.60e-4) = & 9.9906e-1 (1.87e-3) \\
        C1\_DTLZ3 & 10 & 0.0000e+0 (0.00e+0) = & 0.0000e+0 (0.00e+0) & 0.0000e+0 (0.00e+0) = & 0.0000e+0 (0.00e+0) \\ 
        C2\_DTLZ2 & 10 & 1.0909e-1 (5.76e-1) = & 1.0909e-1 (8.48e-8) & 6.7119e-1 (2.76e-3) = & 6.7081e-1 (2.02e-3) \\
        C3\_DTLZ4 & 10 & 9.9950e-1 (3.30e-5) = & 9.9949e-1 (2.10e-5) & 9.9904e-1 (5.20e-5) = & 9.9903e-1 (4.80e-5) \\ 
        DC1\_DTLZ1 & 10 & 9.9950e-1 (1.21e-2) = & 9.9954e-1 (7.47e-4) & 9.5178e-1 (8.30e-2) = & 9.7549e-1 (5.96e-2) \\ 
        DC1\_DTLZ3 & 10 & 9.6580e-1 (3.50e-4) = & 9.6582e-1 (3.09e-4) & 6.4555e-1 (2.56e-1) = & 6.6735e-1 (2.36e-1) \\ 
        DC2\_DTLZ1 & 10 & 9.6942e-1 (3.59e-2) = & 9.7162e-1 (3.39e-2) & 9.9953e-1 (1.05e-4) = & 9.9955e-1 (1.01e-4) \\ 
        DC2\_DTLZ3 & 10 & 0.0000e+0 (0.00e+0) = & 0.0000e+0 (0.00e+0) & 9.2501e-1 (8.95e-1) = & 9.0257e-1 (6.32e-1) \\
        DC3\_DTLZ1 & 10 & 3.4328e-1 (9.24e-3) = & 3.4386e-1 (6.94e-3) & 0.0000e+0 (0.00e+0) = & 0.0000e+0 (0.00e+0) \\
        DC3\_DTLZ3 & 10 & 0.0000e+0 (0.00e+0) = & 0.0000e+0 (0.00e+0) & 6.5739e-1 (9.40e-2) = & 6.7940e-1 (8.49e-2) \\ 
        +/-/= &   & 4/1/35 & ~ & 0/0/40 & ~\\ 
  \bottomrule
\end{tabular}
\end{center}
\end{footnotesize}{$+$, $-$, and $=$ denote the performance of the selected algorithm is significantly better, worse, and equivalent to the corresponding variant, respectively.}
%\end{footnotesize}
\end{table*}

\begin{figure}%[htbp]
  \centering
  \includegraphics[width=\textwidth]{Figures/All/1CNSGAII-DTLZ.pdf}
  \caption{The scatter plots of the non-dominated solution set (denoted as red circles) obtained by \texttt{C-NSGA-II} and $v$\texttt{C-NSGA-II} (dubbed C-DTLZcu and DC-DTLZcu) on C-DTLZ and DC-DTLZ test problems.}
  \label{DTLZ-C-NSGA-II}
\end{figure}

\begin{figure}%[htbp]
  \centering
  \includegraphics[width=\textwidth]{Figures/All/1CNSGAIII-DTLZ.pdf}
  \caption{The scatter plots of the non-dominated solution set (denoted as red circles) obtained by \texttt{C-NSGA-III} and $v$\texttt{C-NSGA-III} (dubbed C-DTLZcu and DC-DTLZcu) on C-DTLZ and DC-DTLZ test problems.}
  \label{DTLZ-C-NSGA-III}
\end{figure}

\begin{figure}%[htbp]
  \centering
  \includegraphics[width=\textwidth]{Figures/All/1CMOEAD-DTLZ.pdf}
  \caption{The scatter plots of the non-dominated solution set (denoted as red circles) obtained by \texttt{C-MOEA/D} and $v$\texttt{C-MOEA/D} (dubbed C-DTLZcu and DC-DTLZcu) on C-DTLZ and DC-DTLZ test problems.}
  \label{DTLZ-C-MOEA/D}
\end{figure}

\begin{figure}%[htbp]
  \centering
  \includegraphics[width=\textwidth]{Figures/All/1CTAEA-DTLZ.pdf}
  \caption{The scatter plots of the non-dominated solution set (denoted as red circles) obtained by \texttt{C-TAEA} and $v$\texttt{C-TAEA} (dubbed C-DTLZcu and DC-DTLZcu) on C-DTLZ and DC-DTLZ test problems.}
  \label{DTLZ-C-TAEA}
\end{figure}

\subsection{Results on Real World Test Problems}
 Since the optimal PFs of RWCMOPs are unknown, we only perform the Wilcoxon signed-rank test on HV metric of it and report the reults in the Table~\ref{tab:RWCMOPsHVa} and Table~\ref{tab:RWCMOPsHVb}. As shown in these tables, the most of comparison results (ranging from 68.5\% to 85.7\%) do not have significant difference by statistics analysis. It means that all four peer algorithms can keep their performance in the majority of RWCMOPs problems when replacing the CV with a crisp value. To better analyse the characteristics of these problems, we draw out the scatter plot of feasible regions by recording the fitness values of both feasible and infeasible individuals producing in the optimization process (best HV metric), the figures are shown in Figure \ref{RCM-C-NSGA-II}, \ref{RCM-C-NSGA-III}, \ref{RCM-C-MOEA/D}, \ref{RCM-C-TAEA}.

   \begin{table*}%[htbp]
   \begin{footnotesize}
   \begin{center}
  \caption{Comparison results on HV metric (median and IQR) for C-NSGA-II and C-NSGA-III against their corresponding variants on RWCMOPs}
  \label{tab:RWCMOPsHVa}
  \centering
  \begin{tabular}{lllllll}
    \toprule
    Problem & M     & $v$C-NSGA-II & C-NSGA-II & $v$C-NSGA-III & C-NSGA-III \\
    \midrule
    RWMOP1 & 2     & 6.0237e-1 (1.25e-3) = & 6.0240e-1 (8.97e-4) & 6.0618e-1 (1.46e-3) = & 6.0586e-1 (1.44e-3) \\
    RWMOP2 & 2     & 3.1489e-1 (3.17e-1) = & 2.7457e-1 (1.65e-1) & 1.7795e-1 (3.08e-1) = & 2.9169e-1 (2.08e-1) \\
    RWMOP3 & 2     & 9.0143e-1 (2.93e-4) - & 9.0155e-1 (2.79e-4) & 8.9150e-1 (1.06e-3) = & 8.9132e-1 (1.11e-3) \\
    RWMOP4 & 2     & 8.5842e-1 (4.65e-3) = & 8.5902e-1 (5.90e-3) & 8.5398e-1 (1.07e-2) = & 8.5456e-1 (1.35e-2) \\
    RWMOP5 & 2     & 4.3314e-1 (1.35e-3) = & 4.3322e-1 (2.14e-3) & 4.3314e-1 (1.29e-3) = & 4.3291e-1 (2.05e-3) \\
    RWMOP6 & 2     & 2.7711e-1 (4.42e-5) = & 2.7712e-1 (3.17e-5) & 2.7663e-1 (2.59e-4) = & 2.7666e-1 (2.85e-4) \\
    RWMOP7 & 2     & 4.8354e-1 (1.89e-4) = & 4.8353e-1 (1.49e-4) & 4.8364e-1 (1.23e-3) = & 4.8374e-1 (3.43e-4) \\
    RWMOP8 & 3     & 2.5947e-2 (1.08e-4) = & 2.5934e-2 (9.76e-5) & 2.5638e-2 (1.68e-4) = & 2.5591e-2 (1.88e-4) \\
    RWMOP9 & 2     & 4.0808e-1 (3.21e-4) = & 4.0810e-1 (1.73e-4) & 4.0953e-1 (1.40e-4) = & 4.0952e-1 (1.08e-4) \\
    RWMOP10 & 2     & 8.4720e-1 (3.01e-4) = & 8.4712e-1 (3.28e-4) & 8.3310e-1 (3.55e-5) = & 8.3310e-1 (9.63e-5) \\
    RWMOP11 & 5     & 1.0040e-1 (5.15e-4) = & 1.0026e-1 (7.87e-4) & 1.0007e-1 (6.33e-4) = & 1.0005e-1 (4.13e-4) \\
    RWMOP12 & 2     & 5.5868e-1 (5.59e-4) = & 5.5871e-1 (9.40e-4) & 5.5939e-1 (1.64e-3) = & 5.5957e-1 (1.75e-3) \\
    RWMOP13 & 3     & 8.9298e-2 (2.54e-4) - & 8.9384e-2 (2.71e-4) & 9.0014e-2 (3.70e-4) = & 9.0095e-2 (2.11e-4) \\
    RWMOP14 & 2     & 6.1832e-1 (8.91e-4) = & 6.1841e-1 (8.34e-4) & 6.1798e-1 (9.90e-4) = & 6.1819e-1 (1.34e-3) \\
    RWMOP15 & 2     & 5.0863e-1 (1.22e-2) - & 5.4159e-1 (7.08e-4) & 4.9561e-1 (3.18e-2) - & 5.4100e-1 (1.33e-3) \\
    RWMOP16 & 2     & 7.6303e-1 (2.47e-4) = & 7.6299e-1 (2.59e-4) & 7.6240e-1 (8.55e-5) = & 7.6240e-1 (7.75e-5) \\
    RWMOP17 & 3     & 2.6712e-1 (8.20e-3) = & 2.6925e-1 (6.89e-3) & 2.6258e-1 (1.62e-2) = & 2.6218e-1 (9.20e-3) \\
    RWMOP18 & 2     & 4.0461e-2 (1.28e-5) - & 4.0464e-2 (9.09e-6) & 4.0500e-2 (6.77e-6) - & 4.0505e-2 (4.78e-6) \\
    RWMOP19 & 3     & 3.3173e-1 (1.23e-2) - & 3.3730e-1 (1.10e-2) & 3.2255e-1 (1.28e-2) = & 3.2205e-1 (1.23e-2) \\
    RWMOP20 & 2     & 0.0000e+0 (0.00e+0) = & 0.0000e+0 (0.00e+0) & 0.0000e+0 (0.00e+0) = & 0.0000e+0 (0.00e+0) \\
    RWMOP21 & 2     & 3.1744e-2 (1.61e-5) = & 3.1744e-2 (5.91e-6) & 3.1730e-2 (8.29e-5) = & 3.1738e-2 (6.45e-5) \\
    RWMOP22 & 2     & 0.0000e+0 (0.00e+0) = & 0.0000e+0 (0.00e+0) & 0.0000e+0 (0.00e+0) = & 0.0000e+0 (0.00e+0) \\
    RWMOP23 & 2     & 0.0000e+0 (0.00e+0) - & 0.0000e+0 (3.27e-1) & 0.0000e+0 (0.00e+0) - & 1.3726e-1 (2.78e-1) \\
    RWMOP24 & 3     & 0.0000e+0 (0.00e+0) = & 0.0000e+0 (0.00e+0) & 0.0000e+0 (0.00e+0) = & 0.0000e+0 (0.00e+0) \\
    RWMOP25 & 2     & 2.4055e-1 (1.68e-4) = & 2.4059e-1 (1.67e-4) & 2.4106e-1 (1.96e-5) = & 2.4106e-1 (2.53e-5) \\
    RWMOP26 & 2     & 1.3296e-1 (3.94e-2) - & 1.4385e-1 (2.59e-2) & 1.1004e-1 (6.09e-2) - & 1.3640e-1 (4.51e-2) \\
    RWMOP27 & 2     & 1.6437e+9 (2.89e+9) = & 1.6523e+9 (1.10e+10) & 8.6553e+5 (3.77e+5) = & 9.7120e+5 (3.72e+5) \\
    RWMOP28 & 2     & 0.0000e+0 (0.00e+0) = & 0.0000e+0 (0.00e+0) & 0.0000e+0 (0.00e+0) = & 0.0000e+0 (0.00e+0) \\
    RWMOP29 & 2     & 7.8143e-1 (2.00e-2) = & 7.5768e-1 (3.24e-2) & 7.8232e-1 (2.40e-2) = & 7.8192e-1 (2.71e-2) \\
    RWMOP30 & 2     & 0.0000e+0 (0.00e+0) - & 5.2368e-1 (6.29e-1) & 0.0000e+0 (0.00e+0) - & 4.7596e-1 (6.39e-1) \\
    RWMOP31 & 2     & 0.0000e+0 (0.00e+0) - & 8.0907e-2 (3.86e-1) & 0.0000e+0 (0.00e+0) - & 1.1639e-1 (5.63e-1) \\
    RWMOP32 & 2     & 0.0000e+0 (0.00e+0) - & 6.7551e-1 (7.81e-1) & 0.0000e+0 (0.00e+0) - & 5.9671e-1 (7.30e-1) \\
    RWMOP33 & 2     & 0.0000e+0 (0.00e+0) = & 0.0000e+0 (0.00e+0) & 0.0000e+0 (0.00e+0) = & 0.0000e+0 (0.00e+0) \\
    RWMOP34 & 2     & 0.0000e+0 (0.00e+0) = & 0.0000e+0 (0.00e+0) & 0.0000e+0 (0.00e+0) = & 0.0000e+0 (0.00e+0) \\
    RWMOP35 & 2     & 0.0000e+0 (0.00e+0) - & 3.8710e-1 (5.66e-1) & 0.0000e+0 (0.00e+0) - & 1.9355e-1 (0.00e+0) \\
    +/-/= &   & 0/11/24 & ~ & 0/8/27 & ~\\ 
    %\multicolumn{2}{|c|}{+/-/=} & 0/8/32 &       & 0/15/25 &  \\
  \bottomrule
\end{tabular}
\end{center}
\end{footnotesize}{$+$, $-$, and $=$ denote the performance of the selected algorithm is significantly better, worse, and equivalent to the corresponding variant, respectively.}
\end{table*}

  \begin{table*}%[htbp]
   \begin{footnotesize}
   \begin{center}
  \caption{Comparison results on HV metric (median and IQR) for C-MOEA/D and C-TAEA against their corresponding variants on RWCMOPs}
  \label{tab:RWCMOPsHVb}
  \centering
  \begin{tabular}{lllllll}
    \toprule
    Problem & M     & $v$C-MOEA/D & C-MOEA/D & $v$C-TAEA & C-TAEA\\
    \midrule
    RWMOP1 & 2     & 1.0896e-1 (8.53e-5) = & 1.0897e-1 (8.94e-5) & 6.0384e-1 (1.67e-3) = & 6.0401e-1 (1.63e-3) \\
    RWMOP2 & 2     & 3.6848e-1 (9.30e-2) + & 2.8202e-1 (1.05e-1) & 1.0094e-1 (3.77e-1) = & 1.2832e-1 (2.28e-1) \\
    RWMOP3 & 2     & 1.3838e-1 (9.29e-2) + & 9.4748e-2 (3.20e-2) & 8.8155e-1 (1.51e-2) + & 8.6640e-1 (3.29e-2) \\
    RWMOP4 & 2     & 0.0000e+0 (0.00e+0) = & 0.0000e+0 (0.00e+0) & 8.4567e-1 (2.17e-2) = & 8.4504e-1 (1.70e-2) \\
    RWMOP5 & 2     & 4.2334e-1 (9.22e-3) = & 4.1759e-1 (1.25e-2) & 4.2755e-1 (7.97e-3) = & 4.2913e-1 (8.44e-3) \\
    RWMOP6 & 2     & 2.7636e-1 (3.88e-5) = & 2.7636e-1 (4.95e-5) & 2.5509e-1 (5.05e-2) = & 2.3671e-1 (6.71e-2) \\
    RWMOP7 & 2     & 4.8035e-1 (1.42e-3) = & 4.8011e-1 (1.83e-3) & 4.8337e-1 (6.54e-4) = & 4.8306e-1 (1.04e-3) \\
    RWMOP8 & 3     & 1.0009e-2 (1.10e-3) = & 9.8807e-3 (1.37e-3) & 2.6085e-2 (1.16e-4) = & 2.6069e-2 (1.21e-4) \\
    RWMOP9 & 2     & 5.3055e-2 (5.87e-5) = & 5.3050e-2 (4.77e-5) & 4.0774e-1 (1.24e-3) = & 4.0791e-1 (1.25e-3) \\
    RWMOP10 & 2     & 7.9329e-2 (7.91e-4) = & 7.9451e-2 (7.06e-4) & 8.4093e-1 (7.72e-3) = & 8.4262e-1 (4.17e-3) \\
    RWMOP11 & 5     & 6.0542e-2 (5.23e-4) = & 6.0532e-2 (6.19e-4) & 1.0120e-1 (3.45e-4) = & 1.0122e-1 (5.56e-4) \\
    RWMOP12 & 2     & 7.0197e-2 (2.10e-2) = & 6.8607e-2 (2.38e-2) & 5.4338e-1 (2.04e-2) = & 5.4447e-1 (1.03e-2) \\
    RWMOP13 & 3     & 8.9919e-2 (1.70e-4) - & 9.0099e-2 (2.50e-4) & 8.7668e-2 (2.03e-3) = & 8.8028e-2 (3.24e-3) \\
    RWMOP14 & 2     & 1.3844e-1 (4.26e-2) = & 1.3859e-1 (2.19e-2) & 6.1802e-1 (9.19e-4) = & 6.1777e-1 (1.02e-3) \\
    RWMOP15 & 2     & 4.5545e-1 (6.42e-2) + & 6.6015e-2 (9.59e-7) & 3.0303e-1 (2.25e-1) - & 5.3940e-1 (3.75e-3) \\
    RWMOP16 & 2     & 7.9107e-2 (1.55e-4) = & 7.9071e-2 (7.84e-5) & 7.6136e-1 (8.77e-4) = & 7.6167e-1 (8.09e-4) \\
    RWMOP17 & 3     & 2.5876e-1 (1.02e-1) = & 2.3142e-1 (7.69e-2) & 2.5880e-1 (1.45e-2) = & 2.5844e-1 (2.31e-2) \\
    RWMOP18 & 2     & 4.0255e-2 (3.50e-5) = & 4.0259e-2 (1.91e-5) & 4.0284e-2 (1.26e-4) = & 4.0257e-2 (1.46e-4) \\
    RWMOP19 & 3     & 2.2752e-1 (3.18e-2) + & 1.7718e-1 (6.87e-2) & 1.4541e-1 (2.88e-2) = & 1.3178e-1 (6.58e-2) \\
    RWMOP20 & 2     & 0.0000e+0 (0.00e+0) = & 0.0000e+0 (0.00e+0) & 0.0000e+0 (0.00e+0) = & 0.0000e+0 (0.00e+0) \\
    RWMOP21 & 2     & 2.9322e-2 (4.56e-6) = & 2.9322e-2 (3.34e-6) & 3.1619e-2 (8.35e-5) = & 3.1634e-2 (8.02e-5) \\
    RWMOP22 & 2     & 0.0000e+0 (0.00e+0) = & 0.0000e+0 (0.00e+0) & 0.0000e+0 (0.00e+0) = & 0.0000e+0 (0.00e+0) \\
    RWMOP23 & 2     & 0.0000e+0 (0.00e+0) - & 0.0000e+0 (2.04e-1) & 0.0000e+0 (0.00e+0) = & 0.0000e+0 (0.00e+0) \\
    RWMOP24 & 3     & 0.0000e+0 (0.00e+0) = & 0.0000e+0 (0.00e+0) & 0.0000e+0 (0.00e+0) = & 0.0000e+0 (0.00e+0) \\
    RWMOP25 & 2     & 2.3671e-1 (3.60e-4) = & 2.3683e-1 (4.66e-4) & 2.4078e-1 (2.92e-4) = & 2.4063e-1 (3.92e-4) \\
    RWMOP26 & 2     & 1.2190e-1 (4.49e-2) - & 1.6252e-1 (4.92e-2) & 0.0000e+0 (0.00e+0) - & 1.0323e-1 (1.43e-2) \\
    RWMOP27 & 2     & 2.0467e+2 (5.83e-1) = & 2.0458e+2 (5.40e-1) & 1.4793e+2 (2.74e+2) - & 3.4508e+2 (1.55e+3) \\
    RWMOP28 & 2     & 0.0000e+0 (0.00e+0) = & 0.0000e+0 (0.00e+0) & 0.0000e+0 (0.00e+0) = & 0.0000e+0 (0.00e+0) \\
    RWMOP29 & 2     & 7.7984e-1 (2.22e-2) = & 7.7574e-1 (2.98e-2) & 6.7404e-1 (3.83e-2) = & 6.7886e-1 (5.92e-2) \\
    RWMOP30 & 2     & 0.0000e+0 (0.00e+0) - & 0.0000e+0 (6.31e-1) & 0.0000e+0 (0.00e+0) - & 0.0000e+0 (0.00e+0) \\
    RWMOP31 & 2     & 0.0000e+0 (0.00e+0) - & 7.7809e-2 (4.10e-1) & 0.0000e+0 (0.00e+0) = & 0.0000e+0 (0.00e+0) \\
    RWMOP32 & 2     & 0.0000e+0 (0.00e+0) - & 7.1370e-1 (7.75e-1) & 0.0000e+0 (0.00e+0) = & 0.0000e+0 (0.00e+0) \\
    RWMOP33 & 2     & 0.0000e+0 (0.00e+0) = & 0.0000e+0 (0.00e+0) & 0.0000e+0 (0.00e+0) = & 0.0000e+0 (0.00e+0) \\
    RWMOP34 & 2     & 0.0000e+0 (0.00e+0) = & 0.0000e+0 (0.00e+0) & 0.0000e+0 (0.00e+0) = & 0.0000e+0 (0.00e+0) \\
    RWMOP35 & 2     & 0.0000e+0 (0.00e+0) - & 1.6815e-1 (0.00e+0) & 0.0000e+0 (0.00e+0) = & 0.0000e+0 (0.00e+0) \\
    +/-/= &   & 4/7/24 & ~ & 1/4/30 & ~\\ 
    %\multicolumn{2}{|c|}{+/-/=} & 0/8/32 &       & 0/15/25 &  \\
  \bottomrule
\end{tabular}
\end{center}
\end{footnotesize}{$+$, $-$, and $=$ denote the performance of the selected algorithm is significantly better, worse, and equivalent to the corresponding variant, respectively.}
\end{table*}
 
\begin{figure}%[t]
  \centering
	\includegraphics[width=\textwidth]{Figures/All/CNSGAII-RWMOPs-part1cut.pdf}
\end{figure}
\begin{figure}%[htbp]
  \centering
	\includegraphics[width=\textwidth]{Figures/All/CNSGAII-RWMOPs-part2cut.pdf}
  \caption{Distribution of feasible solutions (denoted as blue circles), infeasible solutions (denoted as black circles) and non-dominated solutions (denoted as red circles) of RCWMOP test problems obtained by \texttt{C-NSGA-II} and $v$\texttt{C-NSGA-II} (dubbed RCMcu).}
  \label{RCM-C-NSGA-II}
\end{figure}

\begin{figure}%[htbp]
  \centering
	\includegraphics[width=\textwidth]{Figures/All/CNSGAIII-RWMOPs-part1cut.pdf}
\end{figure}
\begin{figure}%[htbp]
  \centering
	\includegraphics[width=\textwidth]{Figures/All/CNSGAIII-RWMOPs-part2cut.pdf}
  \caption{Distribution of feasible solutions (denoted as blue circles), infeasible solutions (denoted as black circles) and non-dominated solutions (denoted as red circles) of RCWMOP test problems obtained by \texttt{C-NSGA-III} and $v$\texttt{C-NSGA-III} (dubbed RCMcu).}
  \label{RCM-C-NSGA-III}
\end{figure}

\begin{figure}%[htbp]
  \centering
	\includegraphics[width=\textwidth]{Figures/All/CMOEAD-RWMOPs-part1cut.pdf}
\end{figure}
\begin{figure}%[htbp]
  \centering
	\includegraphics[width=\textwidth]{Figures/All/CMOEAD-RWMOPs-part2cut.pdf}
  \caption{Distribution of feasible solutions (denoted as blue circles), infeasible solutions (denoted as black circles) and non-dominated solutions (denoted as red circles) of RCWMOP test problems obtained by \texttt{C-MOEA/D} and $v$\texttt{C-MOEA/D} (dubbed RCMcu).}
  \label{RCM-C-MOEA/D}
\end{figure}

\begin{figure}%[htbp]
  \centering
	\includegraphics[width=\textwidth]{Figures/All/CTAEA-RWMOPs-part1cut.pdf}
\end{figure}
\begin{figure}%[htbp]
  \centering
	\includegraphics[width=\textwidth]{Figures/All/CTAEA-RWMOPs-part2cut.pdf}
  \caption{Distribution of feasible solutions (denoted as blue circles), infeasible solutions (denoted as black circles) and non-dominated solutions (denoted as red circles) of RCWMOP test problems obtained by \texttt{C-TAEA} and $v$\texttt{C-TAEA} (dubbed RCMcu).}
  \label{RCM-C-TAEA}
\end{figure}

\bibliographystyle{splncs04}
\bibliography{cmo}